%% file: main_ieee_whu-road3d.tex
\documentclass[lettersize,journal]{IEEEtran}
\usepackage{amsmath,amsfonts}
\usepackage{algorithmic}
\usepackage{algorithm}
\usepackage{array}
\usepackage[caption=false,font=footnotesize,labelfont=rm,textfont=rm]{subfig}   
\usepackage{textcomp}
\usepackage{stfloats}
\usepackage{url}
\usepackage{verbatim}
\usepackage{graphicx}
\usepackage{cite}
\usepackage{hyperref} 
\usepackage{multirow}
\usepackage{tabularx}
\usepackage{gensymb}
\begin{document}

\title{WHU-Road3D: A dataset and benchmark towards fine-grained road surface mapping}

\author{Xiaoxin Mi, Zhen Cao, Chong Liu, Yuhao Li, Zhen Dong,~\IEEEmembership{Member,~IEEE,} Bisheng Yang, Yang Jia, Shengwu Xiong
\thanks{Xiaoxin Mi and Shengwu Xiong are with Sanya Science and Education Innovation Park, Wuhan University of Technology, Sanya 572000, China; also with School of Artificial Intelligence, Wuhan University of Technology, Wuhan 430070, China (email: xiaoxin.mi@whut.edu.cn, xiongsw@whut.edu.cn).} 
\thanks{Zhen Cao, Zhen Dong, and Bisheng Yang are with the State Key Laboratory of Information Engineering in Surveying, Mapping, and Remote Sensing (LIESMARS), Wuhan University, Wuhan 430079, China (e-mail: zhen.cao@whu.edu.cn, dongzhenwhu@whu.edu.cn, bshyang@whu.edu.cn).}
\thanks{Yuhao Li is with School of Mechanical Engineering, Chongqing Technology and Business University, Chongqing, China (email: liyuhao@whu.edu.cn).}
\thanks{Yang Jia is with Sichuan Highway Planning, Survey, Design and Research Institute Ltd., Chengdu, China (email: jiayang@schdri.com).}
\thanks{Xiaoxin Mi and Zhen Cao contributed equally to this work. Corresponding authors: Zhen Dong.}
}

\markboth{Journal of \LaTeX\ Class Files,~Vol.~xx, No.~x, August~2026}%
{Shell \MakeLowercase{\textit{et al.}}: A Sample Article Using IEEEtran.cls for IEEE Journals}


\maketitle

\begin{abstract}
Accurately extracting and mapping road surface elements is crucial for applications such as high-definition maps, road asset management, and intelligent transportation systems. This study first provides a comprehensive review of the benchmark datasets for extracting and mapping road elements, encompassing overhead-to-perspective data, large-to-local scales, and coarse-to-fine details. Subsequently, we thoroughly review the relevant approaches for extracting road surface elements.
However, we argue that existing datasets still require enhancements in semantic richness, data modality diversity, and scenario complexity.
To bridge these gaps, this paper presents WHU-Road3D, a large-scale multimodal dataset for fine-grained road surface mapping to address these shortcomings. WHU-Road3D provides approximately 100 km of diverse Chinese highways and urban streets. It features a rich combination of MLS point clouds, associated panoramic images, vehicle trajectories, detailed vectorized annotations, and pointwise instance annotations for road surface elements. The dataset carefully annotates 15 distinct categories of road elements, including common markings like road boundaries, zebra crossings, lane dividers, stop lines, as well as 10 types of direction symbols. To our knowledge, WHU-Road3D stands as the first multimodal dataset for mapping such a comprehensive range of road surface elements, designed to facilitate cutting-edge research in instance segmentation, full-element extraction and mapping on road surfaces. 
Beyond these immediate applications, WHU-Road3D also provides a valuable source for hierarchical road topology reasoning, map updating, and street scene generation.
The dataset, along with the code for our benchmark methods, will be made publicly available at \href{https://github.com/WHU-USI3DV/WHU-Road3D.git}{https://github.com/WHU-USI3DV/WHU-Road3D.git}.
\end{abstract}


\input{sections/1-Introduction}
\input{sections/2-Related_datasets_methods}
\input{sections/3-WHU-Road3D}

\input{sections/4-Benchmarks}	
\input{sections/5-FutureDirections}
\input{sections/6-Conclusion}

\section*{Acknowledgments}
This work was supported by the National Natural Science Foundation of China Projects under Grants No.42401543 and 42130105, the Postdoctor Project of Hubei Province under Grant Number 2024HBBHCXB080, and the Natural Science Foundation of Hainan Province Project under Grants No.625QN390. 
We thank the Shanghai Surveying and Mapping Institute, Shanghai, China, for providing partial source data.  We would also like to thank Jingjie Yan from Wuhan Dynspai Technology Co., Ltd for annotating data in Districts B and C using Point2Model, a point cloud intelligent processing software.

\bibliographystyle{IEEEtran}
\bibliography{rmfile}

\end{document}

%% file: sections/1-Introduction.tex
\section{Introduction}\label{Introduction}
\IEEEPARstart{R}{oad} surfaces mainly comprise asphalt surfaces and various road markings (symbols like arrows, lane dividers, stop lines, and zebra crossings) painted with high-reflective materials, usually bounded by structures called curbs or road boundaries. These road infrastructures regulate the traffic participants' behaviors to build an ordered, efficient, and safe traveling environment. Accurate extraction and mapping of road surface elements are crucial in various applications from past to modern \cite{wongMappingAutonomousDriving2021}, including high-definition maps (HD Maps) for semi-/self-driving vehicles, road infrastructure inventory and management, intelligent transportation systems (ITS), and digital twin cities~\cite{yang2024ubiquitous}.




The last decade witnessed the emergence of deep learning techniques. As for road extraction and mapping, datasets from various sources, views (overhead and perspective), and modalities (images and LiDAR points) have continuously been produced. For instance, RoadTracer~\cite{bastani2018roadtracer} and DeepGlobe~\cite{demirDeepGlobe2018Challenge2018} are from the overhead images to segment roads, while \cite{chengAutomaticRoadDetection2017} further share the road centerline maps alongside the segmentation references. Cityscapes~\cite{Cordts_2016_CVPR} and KITTI360~\cite{kitti360dataset2022} sense the urban roads and sidewalks from perspective images and LiDAR points. \cite{HAN2024500} labels road surface semantics on ALS (Aerial Laser Scanning) and MLS (Mobile Laser Scanning) point clouds. These datasets focus on urban road surface segmentation, where semantic labels are attached to each pixel or point. 
However, pixel-wise or pointwise annotations are difficult to employ for supervising vectorized road mapping tasks, such as HD map generation. Fortunately, nuScenes \cite{caesarNuScenesMultimodalDataset2020} and Argoverse \cite{changArgoverse3DTracking2019, wilsonArgoverse2Next2021} release the map annotations, composing vectorized lane dividers, road boundaries, crosswalks, and their topology relations \cite{wangOpenLaneV2TopologyReasoning2023}, which contribute a lot for the nowadays achievements in advanced urban understanding and HD map construction tasks. 
However, regarding static layers of HD maps and fundamental road infrastructure inventory, the two datasets fail to satisfy the requirements for the diverse scene coverage, fine-grained road surface annotations, and data modality diversity \cite{wongMappingAutonomousDriving2021}. 
\par
To seal these gaps, we are prepared to release the WHU-Road3D, a multimodal dataset for detailed road surface element extraction and mapping. It involves surveying MLS point clouds, panoramic images, associated vehicle trajectories, vectorized annotations, and point-wise instance annotations of road surface elements. 
WHU-Road3D intends to be the key supplementary dataset for road surface object extraction and mapping. It features diverse source data and scenario types, fine-grained vectorized and point-wise annotations.
Specifically, the WHU-Road3D dataset covers typical highway and urban streets in China with 15 categories of road elements, including road boundaries, direction symbols, crosswalks, and lane dividers. Hopefully, WHU-Road3D will serve as a valuable baseline for the relevant research. 

Our main findings are as follows:
\begin{itemize}
\item we comprehensively review the relevant benchmark datasets for road extraction and mapping in three levels of detail, including LoD0 (road centerline), LoD1 (road surface with width), and LoD2 (lanes and direction symbols within the lane).

\item we carefully prepare the WHU-Road3D and relevant baseline methods to continuously enrich the road surface element extraction and mapping datasets, from data modality difference, annotation content richness, and coverage scenario's complexness. 

\item we discover the detailed sub-lane level vectorized annotation and multimodal data can support further explorations, including full-element road surface mapping, spatial layouts assisted mapping, few-shot and open-vocabulary road mapping, hierarchical road topology reasoning, road map updating, lane-level change detection and update, and street vectorized map to scene generation.    
\end{itemize}

The remainder of the paper is organized as follows. Section \ref{Related approaches} first surveys the publicly available datasets for road extraction of three levels of detail, then discusses the relevant studies on LoD2 road surface element extraction and mapping. Section \ref{WHU-Road3D dataset} elaborates on the produced large-scale, fine-grained road surface mapping dataset, WHU-Road3D. Section \ref{sec: benchmark tasks and metrics} presents evaluation metrics, representative benchmark methods, and their performance on WHU-Road3D. Section \ref{Future directions} discusses the challenges and potential prospects of WHU-Road3D.

%% file: sections/2-Related_datasets_methods.tex
\section{Related works}\label{Related approaches}
Roads are the basic infrastructure in the built environment \cite{petersGeographicalPointCloud2018}. Road information extraction and mapping have gained significant interest due to their tremendous downstream applications in road asset inventory and management, reconstruction and expansion, safety evaluation, and (HD) map construction for self-driving cars. Analogous to (3D) building models \cite{grogerOGCCityGeography2012}, road models can also be divided into different levels of detail (LoDs). The higher level often contains more information about roads and requires more effort, but it is not always suitable for applications \cite{petersAutomated3DReconstruction2022}. Therefore, we summarise the road models into three levels of detail, from LoD0 to LoD2. The LoD0 road model represents the road by its centerlines/skeletons. The LoD1 road extends the LoD0 by adding the width information through segmenting road surfaces, extracting road boundaries, etc. Furthermore, LoD2 roads further enrich the road surface information by extending the lane-level details on the road surface, such as lane dividers, direction symbols, and crosswalks.
\par
In the following part, we first survey the existing datasets for road extraction from three levels of detail, then summarize methods for LoD2 road surface mapping.

\subsection{Existing datasets for road extraction}\label{Section_2_benchmark datasets}
Table \ref{table: benchmark datasets} outlines datasets with different observation views, sensors, coverage scenarios, and various annotation details, stimulating road extraction and mapping. 
\begin{table*}[!h]
    \caption {Publicly available datasets for road extraction from the LoD0 to LoD2. When the exact coverage area or road length is not indicated in the original papers, we list the number of cities from which the data were collected.}
    \label{table: benchmark datasets}
	\centering{
	\renewcommand{\arraystretch}{1.3}		
	  \resizebox{0.99\textwidth}{!}
    {%
	\begin{tabular}{llllll}
	\cline{1-6}
	LoD  & Dataset and Reference & Scene & View and sensor & Coverage & Concerned Category \\
    \cline{1-6}
      \multirow{4}{*}{LoD0} & CasNet~\cite{chengAutomaticRoadDetection2017}  & urban area & satellite imagery & 112 km$^{2}$ & road regions \\
      & RoadTracer~\cite{bastani2018roadtracer} & urban area & satellite imagery & 15 cities & road centerline \\  
      & SpaceNet~\cite{ettenSpaceNetRemoteSensing2019}& urban area & satellite imagery & 8676.6 km & road centerline \\
      & Grab-posis~\cite{grab-posisi2019}& urban area & GPS trajectory & 1.0*10$^{6}$ km &  road map inference, traffic forecast\\
      & Sat2Graph~\cite{heSat2GraphRoadGraph2020}& urban area & satellite imagery & 20 cities & road centerline \\
      & MUNO21~\cite{bastaniRoadExtractionDataset2021}& urban area & satellite imagery & 21 cities & road centerline update \\
    \cline{1-6}  
	 \multirow{12}{*}{LoD1} & ISPRS 2D semantic benchmark\footnote{https://www2.isprs.org/commissions/comm2/icwg-2-1a/benchmark\_main/} & urban area & airborne imagery & 2 cities & Impervious surfaces \\  
      & Massachusetts Roads dataset \cite{mnihMachineLearningAerial2013}  & urban area & aerial imagery & 2.634*10$^{3}$ km$^{2}$ & road surfaces \\ 
      \multirow{3}{*}{} & \multirow{3}{*}{TorontoCity dataset \cite{wangTorontoCitySeeingWorld2017}} &  \multirow{3}{*}{urban street} & \multirow{3}{*}{\begin{tabular}[c]{@{}l@{}} aerial and street imagery \\ and LiDAR point cloud, \\ GoPro, stereo image \end{tabular}}  & \multirow{3}{*}{712 km} & \multirow{3}{*}{road surfaces, centerlines, boundaries} \\
      \\
      \\
      & DeepGlobe Road~\cite{demirDeepGlobe2018Challenge2018} & urban, rural area & satellite imagery & 2.22*10$^{3}$ km$^{2}$ & road surfaces \\
      & RoadNet~\cite{liuRoadNetLearningComprehensively2019} & urban area & satellite imagery & 8 km$^{2}$ & road surfaces, centerlines, boundaries \\    
      & \multirow{2}{*}{\cite{sunLeveragingCrowdsourcedGPS2019}} & \multirow{2}{*}{urban area} &  \multirow{2}{*}{\begin{tabular}[c]{@{}l@{}}satellite imagery,\\ GPS trajectory \end{tabular}} & \multirow{2}{*}{~100 km$^{2}$} & \multirow{2}{*}{road surfaces} \\
      \\
      \multirow{2}{*}{} & \multirow{2}{*}{CHN6-CUG~\cite{zhuGlobalContextawareBatchindependent2021}} & \multirow{2}{*}{\begin{tabular}[c]{@{}l@{}} railways, highways, \\ urban and rural roads \end{tabular}} & \multirow{2}{*}{satellite imagery} & \multirow{2}{*}{297.726 km$^{2}$} & \multirow{2}{*}{road surfaces} \\ 
      \\
      & Topo-boundary~\cite{xuTopoBoundaryBenchmarkDataset2021} & urban area & satellite imagery & 5.843 km$^{2}$ & road surfaces and boundaries \\ 
      & GRSet~\cite{luGlobalRoadExtraction2024} & urban area & satellite imagery & 4.95*10$^{4}$ km$^{2}$ & road surfaces \\
      \cline{1-6} 
      \multirow{14}{*}{LoD2} & Caltech Lanes~\cite{alyRealTimeDetection2008} & urban street & front view image & 1225 frames &  lane divider\\
       \multirow{2}{*}{}& \multirow{2}{*}{KITTI road~\cite{geigerAreWeReady2012}}& \multirow{2}{*}{urban street} & \multirow{2}{*}{\begin{tabular}[c]{@{}l@{}}left \& right views of image, \\ Velodyne laser point\end{tabular}}  & \multirow{2}{*}{579 frames} &  \multirow{2}{*}{road surface, lane divider}\\
       \\
       & VPGNet~\cite{leeVPGNetVanishingPoint2017} & urban street & front view image & 2.0*10$^{4}$ frames & 17 types of lane and road marking \\
        & TuSimple~\cite{wangTorontoCitySeeingWorld2017} & highway & front view image & 6048 frames & lane divider \\
        & CULane~\cite{panSpatialDeepSpatial2018} & urban street  & front view image & 133,235 frames & lane divider \\
        & 3DLaneNet~\cite{garnett3DLaneNetEndtoEnd3D2019}& urban street & front view image & 8.5*10$^{4}$ frames & 3D lane divider \\
        & CurveLanes~\cite{xuCurveLaneNASUnifyingLaneSensitive2020} & urban street & front view image & 1.5*10$^{5}$ frames & lane divider \\
        \multirow{2}{*}{}& \multirow{2}{*}{ nuScenes~\cite{caesarNuScenesMultimodalDataset2020}} & \multirow{2}{*}{\begin{tabular}[c]{@{}l@{}}urban street in \\ Boston and Singapore\end{tabular}} &  \multirow{2}{*}{\begin{tabular}[c]{@{}l@{}} LiDAR points, \\ 6 views of image\end{tabular}} & \multirow{2}{*}{\begin{tabular}[c]{@{}l@{}}1000 scenes of \\ 20 second each\end{tabular}}& \multirow{2}{*}{\begin{tabular}[c]{@{}l@{}} road surface, road boundary, \\ crosswalk, lane divider, stop line\end{tabular}} \\
        \\
        & VIL-100~\cite{zhangVIL100NewDataset2021} & urban street & front video & 1.0*10$^{4}$ frames & lane divider \\
        \multirow{2}{*}{}& \multirow{2}{*}{Argoverse V2~\cite{wilsonArgoverse2Next2021}}& \multirow{2}{*}{\begin{tabular}[c]{@{}l@{}}urban street in \\ Miami and Pittsburgh\end{tabular}}  & \multirow{2}{*}{\begin{tabular}[c]{@{}l@{}}LiDAR points, stereo and \\ ring camera imagery\end{tabular}} & \multirow{2}{*}{290 km}  & \multirow{2}{*}{\begin{tabular}[c]{@{}l@{}} lane divider, road boundary, \\ crosswalk\end{tabular}}\\
        \\
        \multirow{2}{*}{}& \multirow{2}{*}{ONCE3DLanes~\cite{yanONCE3DLanesBuildingMonocular2022}}& \multirow{2}{*}{\begin{tabular}[c]{@{}l@{}}highway, bridge, tunnel, \\suburb and downtown\end{tabular}} & \multirow{2}{*}{front view image}& \multirow{2}{*}{2.11*10$^{5}$ frames} & \multirow{2}{*}{3D lane divider}  \\ 
        \\
        \multirow{2}{*}{} & \multirow{2}{*}{OpenLane \cite{chenPersFormer3DLane2022, wangOpenLaneV2TopologyReasoning2023}} & \multirow{2}{*}{urban street} & \multirow{2}{*}{6 or 7 views of image} & \multirow{2}{*}{\begin{tabular}[c]{@{}l@{}}2000 scenes from \\ nuScenes and Argoverse \end{tabular}} & \multirow{2}{*}{\begin{tabular}[c]{@{}l@{}}lane divider, road boundary, crosswalk, \\ lane centerlines and their connections\end{tabular} }\\
        \\
        \multirow{4}{*}{}& \multirow{4}{*}{\textbf{WHU-Road3D}} & \multirow{4}{*}{\textbf{\begin{tabular}[c]{@{}l@{}}urban street, expressway \\ in China\end{tabular}}} & \multirow{4}{*}{\begin{tabular}[c]{@{}l@{}} \textbf{LiDAR points,} \\ \textbf{panoramic image,} \\ \textbf{associated trajectory}\end{tabular}} & \multirow{4}{*}{\textbf{98 km}} & \multirow{4}{*}{\begin{tabular}[c]{@{}l@{}}\textbf{3D road boundary}, \\ \textbf{3D lane divider (dashed, solid)}, \\ \textbf{3D crosswalk, 3D stop line}, \\ \textbf{10 types of 3D direction symbol}\end{tabular}}\\
        \\
        \\
        \\
	\cline{1-6}              
	\end{tabular}%
	}
    }
\end{table*}

\subsubsection{LoD0 road datasets}
The LoD0 road map (road centerline) is crucial for accessibility analysis and crisis response. Several datasets containing aerial images or GPS trajectories have been released for LoD0 road extraction.
Notable datasets include RoadTracer~\cite{bastani2018roadtracer} and Sat2Graph~\cite{heSat2GraphRoadGraph2020}. These datasets are labeled manually or modified from the coarse annotations derived from OpenStreetMap (OSM)~\cite{haklayOpenStreetMapUserGeneratedStreet2008}. However, due to the resolution limitations of the source aerial images, the precision of annotated road centerlines is often suboptimal.
Also, coverage area, public availability, and annotation style limit the potential for foundational road mapping. Therefore, SpaceNet~\cite{ettenSpaceNetRemoteSensing2019} labels the road centerlines on satellite images in four cities, Vegas, Paris, Shanghai, and Khartoum, as well as four other attributes: road type, surface type, bridge, and lane number, to support the research about the large-scale road centerline extraction.

\cite{grab-posisi2019} summarize publicly available GPS trajectory datasets collected from smartphones or dedicated GPS devices worldwide. The quality of the datasets varies in data noise, sampling rate, dataset geography, etc. Those noisy GPS trajectories can be transformed into geographical road networks \cite{CellNet2018, RoadExtractionGaoLipeng2021, DuARE2022}.   
Apart from road extraction,~\cite{bastaniRoadExtractionDataset2021} develop the MUNO21 dataset for the map update task, in which a time series of aerial images are included, constructed roads, deconstructed roads, and road network topology changes are annotated in the form of road centerlines as well. They argue that more attention is required to update existing maps by adding, removing, and shifting roads without introducing errors to keep the map up-to-date rather than infer new ones.

\subsubsection{LoD1 road datasets}
Besides road centerlines, LoD1 road datasets care about road regions and their boundaries. 
The ISPRS 2D semantic labeling benchmark\footnote{https://www2.isprs.org/commissions/comm2/icwg-2-1a/benchmark\_main/}, Massachusetts Roads dataset \cite{mnihMachineLearningAerial2013}, CasNet~\cite{chengAutomaticRoadDetection2017}, DeepGlobe~\cite{demirDeepGlobe2018Challenge2018} and RoadNet~\cite{liuRoadNetLearningComprehensively2019} are pioneering overhead aerial imagery datasets for foundational urban and rural road mapping.
Roads in mountainous areas usually present different characteristics. Therefore, \cite{zhouSplitDepthWiseSeparable2022} and DualNet-PoiD~\cite{zhangDualNetPoiDHybridNeural2024} introduce LoD1 roads dataset covering mountainous areas.
To enhance model generalization, new datasets like CHN6-CUG~\cite{zhuGlobalContextawareBatchindependent2021} and GRSet~\cite{luGlobalRoadExtraction2024} offer significantly more extensive scale and diversity.
Besides road segmentation, Topo-boundary~\cite{xuTopoBoundaryBenchmarkDataset2021} introduced the first dataset for extracting road boundaries from aerial images, comprising 25k annotated images and using polylines to model road boundary lines.

Additionally, to address aerial images' lighting, shadows, and occlusion issues, some researchers build multimodal road extraction datasets using one or several data modalities, including point clouds, crowdsource trajectories, overhead images, and perspective images. \cite{luoDeepCrossModalFusion2024} employs overhead high-resolution images and LiDAR points in Erie Dataset\footnote{https://see-mike-out.github.io/erie-documentation/} and USGS Dataset\footnote{https://data.usgs.gov/datacatalog/data/USGS:ad3d631d-f51f-4b6a-91a3-e617d6a58b4e} to segment road regions.
Those data are from the U.S. Geological Survey\footnote{https://www.usa.gov/agencies/u\-s\-geological\-survey} and Google Maps\footnote{https://developers.google.com/maps/documentation/maps-static/intro}, part of data are not publicly available. 
Other approaches have leveraged aerial imagery and GPS trajectory data to create road extraction datasets~\cite{sunLeveragingCrowdsourcedGPS2019}, which aim to enhance the robustness of road extraction. TorontoCity dataset \cite{wangTorontoCitySeeingWorld2017} contains 8439 $km$ of roads labeled with centerlines and curbs, in which aerial images, street-view panorama, GoPro, stereo image, street-view, and aerial LiDAR point clouds are available.

\subsubsection{LoD2 road datasets}
The LoD2 road map highly enriches the road information by sensing lane-level information, including lane dividers, direction symbols, zebra crossings, stop lines, etc. Perspective images and MLS point clouds collected via the moving vehicle are usually employed to perceive the finer road information. 
At first, massive datasets containing perspective images emerged to advance lane detection techniques, from front view only to multiple perspectives, including Caltech Lanes dataset~\cite{alyRealTimeDetection2008}, VPGNet~\cite{leeVPGNetVanishingPoint2017}, TuSimple~\cite{wangTorontoCitySeeingWorld2017}, CULane~\cite{panSpatialDeepSpatial2018}, 3DLaneNet~\cite{garnett3DLaneNetEndtoEnd3D2019}, CurveLanes~\cite{xuCurveLaneNASUnifyingLaneSensitive2020}, GenLaneNet~\cite{guoGenLaneNetGeneralizedScalable2020}, VIL-100~\cite{zhangVIL100NewDataset2021}, and ONCE3DLanes~\cite{yanONCE3DLanesBuildingMonocular2022}. 
Meanwhile, some researchers noticed LiDAR points can preserve 3D information and reflective intensity. KITTI~\cite{geigerAreWeReady2012}, KLane \cite{paekKLaneLidarLane2022}, and WHU-Lane3D~\cite{miBenchmarkApproachDataset2024} are releasing LiDAR points dataset for lane mapping.
With the rapid development of self-driving vehicles, road elements, such as road boundaries, crosswalks, stop lines, and direction symbols, are needed to improve the safety and stability of autonomous driving systems. Argoverse V2~\cite{wilsonArgoverse2Next2021}, nuScenes~\cite{caesarNuScenesMultimodalDataset2020}, OpenLane~\cite{chenPersFormer3DLane2022}, and OpenLane V2~\cite{wangOpenLaneV2TopologyReasoning2023} are large-scale road scenario understanding and mapping datasets, enabling HD map construction for autonomous driving concerning 3D lanes, road boundaries, and crosswalks. 

\subsection{Road surface element extraction mothods}\label{Section_2_benchmark_methods}
Many studies have attempted to improve the robustness and generalization of the methods for producing results with better continuity, completeness, and correctness \cite{zhuGlobalContextawareBatchindependent2021, daiRADANetRoadAugmented2023, chenSemiRoadExNetSemisupervisedNetwork2023}.
Previous reviews~\cite{WANG2016271, abdollahiDeepLearningApproaches2020, lianRoadExtractionMethods2020} have summarized the road extraction methods from (very) high-resolution images. Apart from overhead source data, \cite{luoRoadObjectDetection2023} provided a comprehensive review of the methods and datasets about road surface information extraction, in which pavement, road markings, and lane dividers extraction methods are summarised. However, methods for road surface element vectorization are missed, which are significant in detailed road surface mapping.
Detailed road surface (LoD2) information extraction methods pay close attention to road details, including road boundaries, lane dividers, directional symbols, crosswalks, stop lines, etc. According to the representation style, the relevant methods can be categorized into unstructured information extraction and structured representation of road elements, respectively. Results in pixel clusters and point clouds are unstructured representations, while polylines and polygons are considered structured representations. Practical applications often prefer structured road representations to pixels/points, for the information in structured representations is well-structured, compact, and explicitly labeled \cite{petersGeographicalPointCloud2018}.
\par
\subsubsection{Unstructured extraction of road information}
    Unstructured extraction of road surface elements typically relies on color and intensity features to segment elements from the background.         
    Previous methods for road surface object extraction primarily relied on images captured by onboard cameras, leveraging pixel-wise hand-craft features and deep learning techniques to separate lane and road markings from the background \cite{panSpatialDeepSpatial2018, li2019line}. However, image-based methods usually struggle with shadows, resolution reduction along camera rays, and lack of accurate 3D geometry.  
    Fortunately, LiDAR data naturally possess three-dimensional spatial information and can reach spatial accuracy at the centimeter level, making it an attractive data source for detailed road information extraction.         
    Segmentation-based methods have been introduced for road alignment extraction \cite{zang2017lane, li2022hdmapnet, li20223d}. Some works are focused on lane connectivity, such as TR \cite{can2022topology}, which introduced minimal cycles and coverings to represent road topology using directed lane curves and their interactions. As for road marking extraction, \cite{yang20173d} introduced the Binary Kernel Descriptor (BKD), a novel 3D localized feature for identifying road marking points. \cite{ghallabi2018lidar} segmented road points, projected them onto an image, and detected lane markings using the Hough transform. \cite{Wen2019178} introduced a deep learning framework for extracting and completing traffic symbols from MLS data, with a completion step to handle missing objects.  \cite{rastiveis2020automated} used geometric and radiometric descriptors to extract lane marking points and developed a fuzzy inference algorithm for pavement lane marking extraction.  
    The above methods extract road information and store it in pixel or point clusters. Although this is crucial for understanding different scenarios, this type of representation leads to data redundancy for storage and downstream applications. Consequently, many researchers focus on structured representation or directly extracting road elements in a structured form from images/point clouds.

\subsubsection{ Structured mapping of road surface elements}      
    According to spatial layout, road surface elements can be categorized into two types: road alignment-related (lane dividers and road boundaries) and traffic symbols (directional symbols, characters, crosswalks, stop lines, etc.). Parametric curves, such as circular, spirals, Bezier, B-spline, polylines, etc., are the structured representation of road alignment-related elements. Meanwhile, traffic symbols are represented via polygons, combinations of location and orientation, outlines, etc.

    Methods for extracting traffic symbols can be broadly classified into top-down and bottom-up approaches. The bottom-up approach identifies the location of road symbols firstly segment interested pixels/points, and then vectorized using some post-processing methods such as morphological operations \cite{JUNG20191}, region growing \cite{cheng2016extraction}, and conditional random fields. \cite{9056457} proposed a finely tuned image-to-image translation model for extracting road markings from feature images, followed by a node averaging and optimized iterative closest point algorithm to generate 3D vectorized representations. Top-down methods, on the other hand, often utilize pre-defined models to guide road marking reconstruction\cite{hervieu2015road,prochazka2019automatic}. However, without prior knowledge of the number and category of road markings, the optimization problem encountered a vast search space. To summarize, bottom-up methods struggle with imperfect raw data, while top-down methods are often time-consuming due to large search spaces. To address these issues, \cite{miTwostageApproachRoad2021a} proposed a two-stage approach for extracting and modeling road markings: first detecting the coarse location of road markings using an object detection network, then refining the location and scale using a template matching algorithm to achieve vectorization.
          
    Methods for extracting road alignment are categorized into ordered points-based and parametric model-based approaches. 
    Ordered points-based methods represent the road alignment as a sequence of sparsely ordered points. \cite{homayounfar2018hierarchical} and \cite{xuTopoBoundaryBenchmarkDataset2021} utilized RNN to trace the points within a road alignment. However, the initialization quality significantly influences the subsequent extraction process, potentially leading to error accumulation. Therefore, line anchors were introduced for lane detection tasks. \cite{garnett3DLaneNetEndtoEnd3D2019} used longitudinal anchors to represent lanes by regressing vertices after extracting lane-related features. However, line-anchor-based approaches struggle with complex lane topologies due to the longitudinal assumption of lane anchors. To address these challenges, 3D LaneNet+ \cite{efrat20203d} and  CondLaneNet \cite{liu2021condlanenet} were proposed to handle lanes with complex topologies.
    As for the parametric model-based methods, Hough Transform \cite{zheng2018improved}, RANSAC \cite{borkar2009robust}, polynomial fitting \cite{neven2018towards} and neural networks\cite{jin2022eigenlanes} are used to fit parametric models from discrete lane pixels or points. For instance, \cite{feng2022rethinking} represented lanes as cubic Bezier curves and developed a deformable convolution-based feature fusion module to explore lane symmetry properties. VectorMapNet~\cite{liu2023vectormapnet}, MapTR~\cite{MapTR}, and MapTRv2~\cite{maptrv2} introduced an end-to-end vectorized high-precision map learning pipeline, formulating the element mapping task as a polyline regression problem. Following MapTR, a series of methods \cite{chen2024maptracker, StreamMapNet2024_WACV, HIMap2024_CVPR, wijaya2024highdefinitionmapmapping} are proposed to improve the HD map generation performance on nuScenes~\cite{caesarNuScenesMultimodalDataset2020}, Argoverse~\cite{wilsonArgoverse2Next2021}, and OpenLane-v2~\cite{wangOpenLaneV2TopologyReasoning2023}.
    \par
    In summary, unstructured mapping methods utilize textural features and require complex post-processing, as discrete, unordered results cannot be directly used. Conversely, structured mapping methods combine a priori knowledge with textural features, considering the geometric characteristics of road surface markings, making them increasingly prevalent in recent research. 
    \par
    However, the above reviews showcase that public benchmark datasets and published studies often focus on road boundaries, lane dividers, and crosswalks. Few public benchmark datasets about 3D direction symbols on road surfaces are available. However, full-element road surface element extraction and mapping are crucial for road asset inventory and maintenance, HD map generation, and map update. Consequently, we cooperate with expert annotators to prepare WHU-Road3D to support research on multimodal full-element road surface element extraction and mapping. 

%% file: sections/3-WHU-Road3D.tex
**\section{WHU-Road3D}\label{WHU-Road3D dataset}

\begin{figure*}[!htbp]
\centering
\subfloat[District-SH]{\includegraphics[width=0.3\textwidth, height=0.32\textwidth] {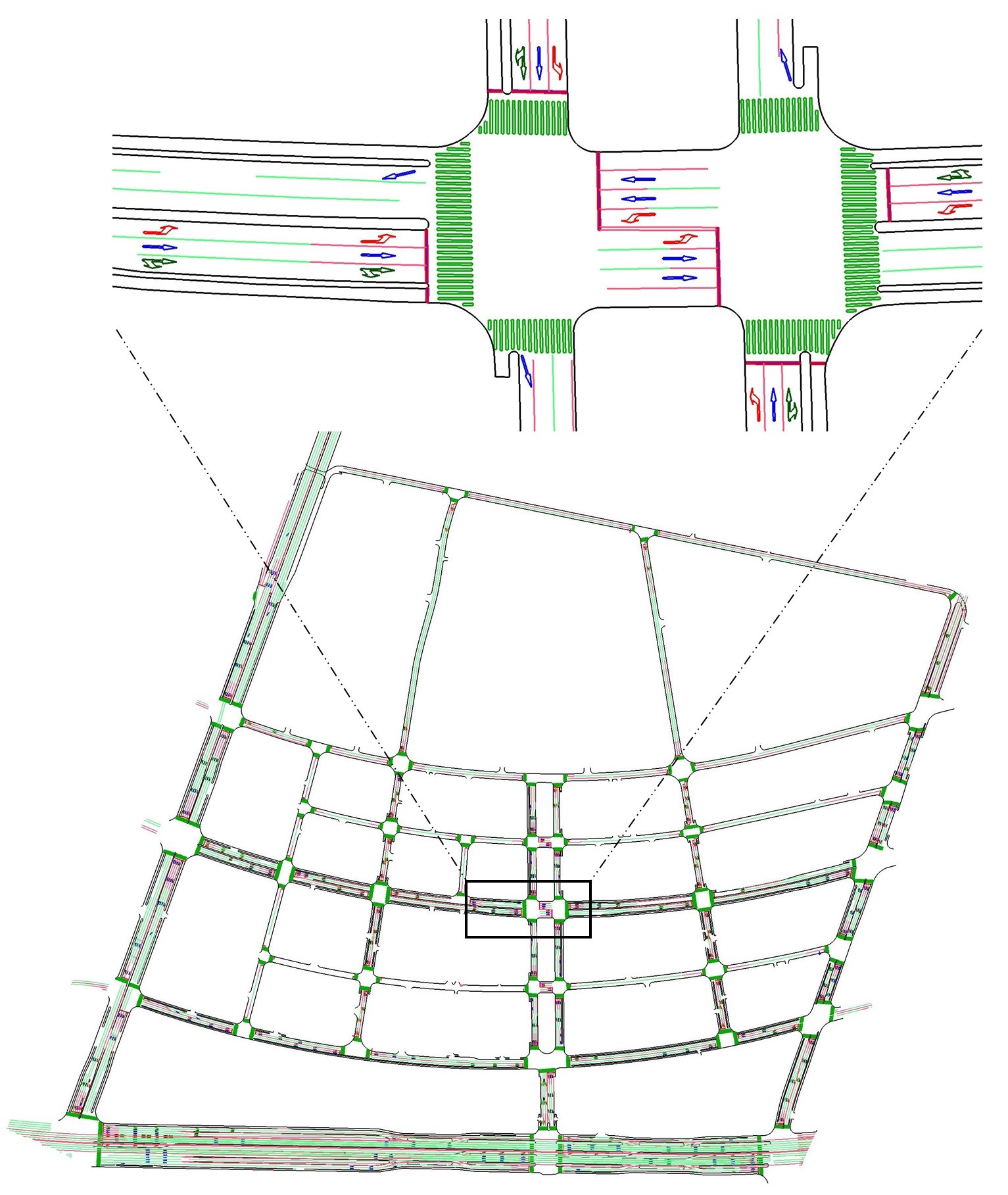}\label{fig_urban_district_A}}
\hfil
\subfloat[District-WH]{\includegraphics[width=0.3\textwidth, height=0.32\textwidth] {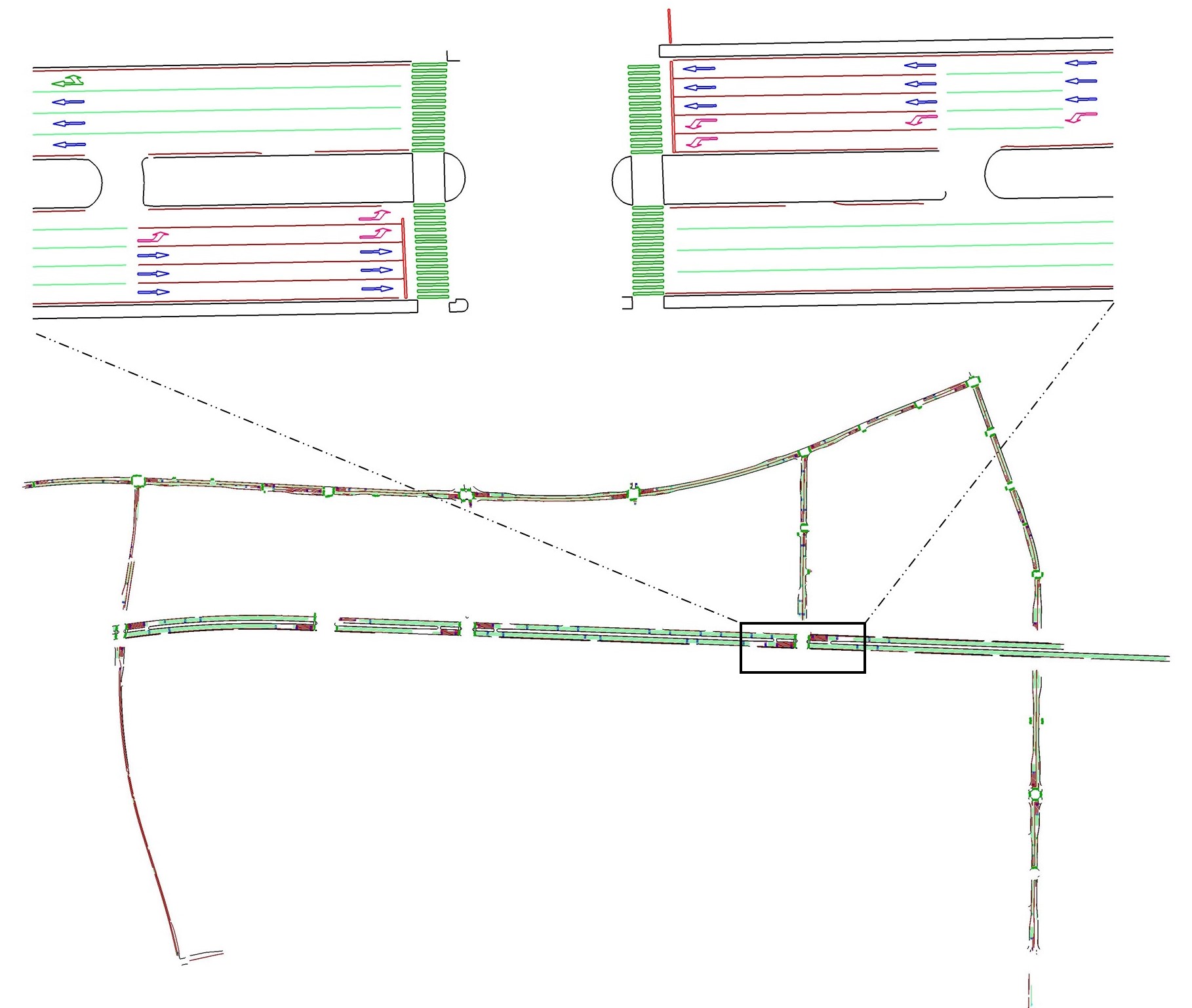}\label{fig_urban_district_B}}
\hfil
\subfloat[District-CM]{\includegraphics[width=0.3\textwidth, height=0.32\textwidth] {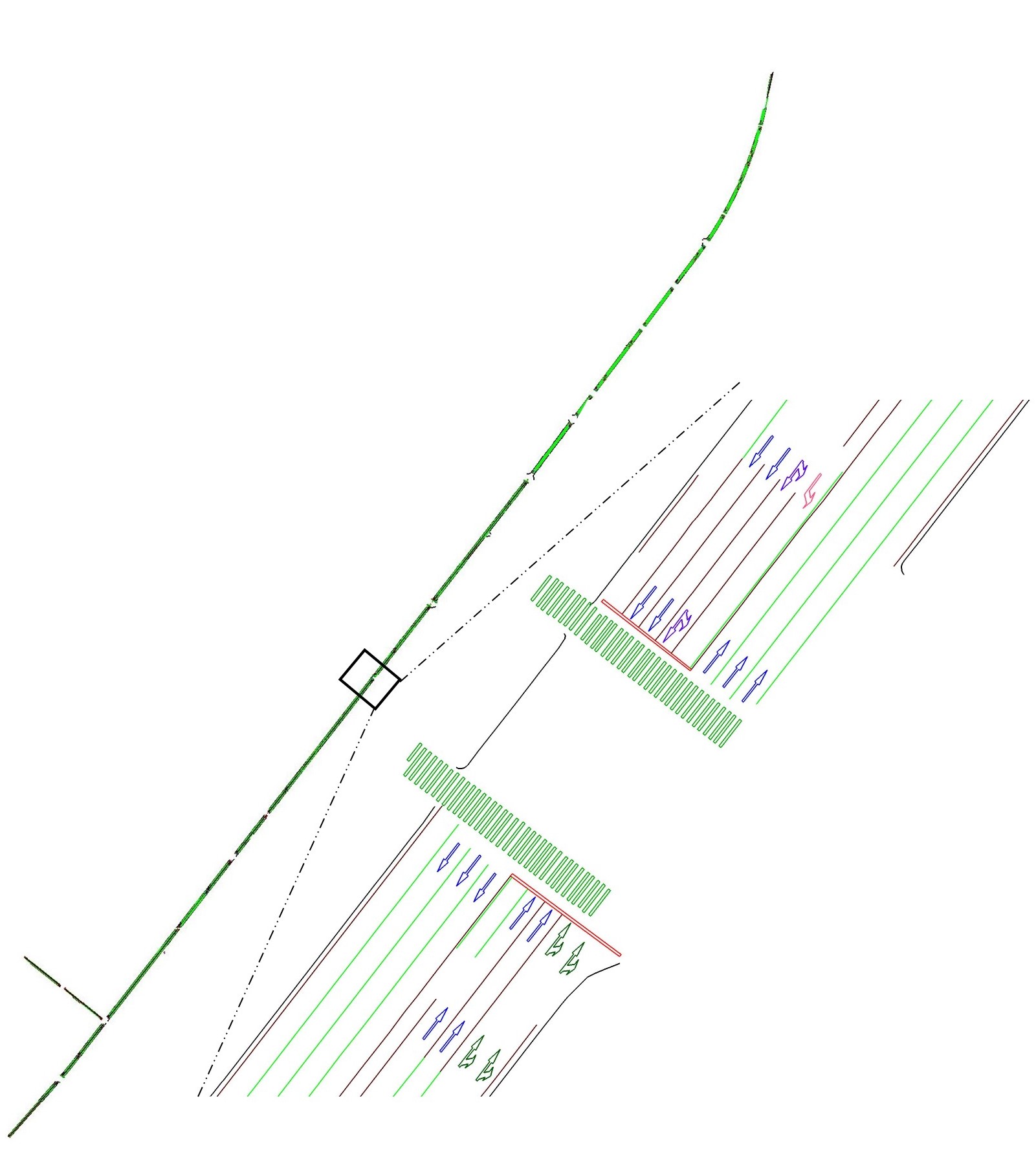}\label{fig_highway_district_A}}
\caption {LoD2 road maps of three districts in WHU-Road3D.}
\label{fig: four districts in WHU-Road3D}
\end{figure*}


\begin{table*}[!htbp]
    \caption {WHU-Road3D statistics in each district.}
    \label{table: dataset description}
	\centering{
	\renewcommand{\arraystretch}{1.3}		
	\resizebox{0.9\textwidth}{!}
    {%
	\begin{tabular}{llllllllll}
	\cline{1-10}
	\multirow{3}{*}{District} & \multirow{3}{*}{\begin{tabular}[c]{@{}l@{}} Road length  \\  (km)\end{tabular}} & \multicolumn{7}{c}{ Annotation statistics } &    \multirow{3}{*}{Scenario}\\
    \cline{3-9} 
     & & \multirow{2}{*}{\begin{tabular}[c]{@{}l@{}} Road boundary  \\  (km)\end{tabular}} & 
     \multicolumn{2}{c}{ Lane divider (km)} & \multicolumn{2}{c}{Crosswalk (count)} &  
     \multirow{2}{*}{\begin{tabular}[c]{@{}l@{}} Stop line\\  (count)\end{tabular}} &
     \multirow{2}{*}{\begin{tabular}[c]{@{}l@{}} Direction symbol  \\  (count)\end{tabular}}  &  \\ 
     \cline{4-7}
     &  &  & Dashed & Solid & Outline & Stripe &  &  \\
	\cline{1-10}
	   SH  & 35.153      & 51.407 &  36.860 & 30.156  & 163 & 2447& 168&  633 & Urban street + expressway \\
      WH  & 26.180      & 23.300 &  22.797 & 34.490 &  88 & 1195& 51&  438 & Urban street   \\
      CM  & 34.185      & 32.782 &  73.779  & 47.178 &  65 & 1350& 71&  511 & Urban street   \\
	\cline{1-10}              
	\end{tabular}%
	}
    }
\end{table*}

\subsection{Data collection}\label{subsec:data collection}
The WHU-Road3D dataset incorporates surveying MLS point clouds from the mobile measurement system, which is equipped with a long-range and precise laser scanner, a high-precision inertial measurement unit (IMU), an advanced global navigation satellite system (GNSS) receiver, and/or a high-resolution high dynamic range (HDR) panoramic camera. Sensors within the system are spatiotemporal synchronized. With the vehicle platform moving, the mobile measurement system collects LiDAR points and panoramic images and records self-pose simultaneously in the street. The collected data covers road boundaries, lane dividers, crosswalks, stop lines, and direction symbols on the ground.
\par
WHU-Road3D consists of LiDAR points on around 98 km of road covering urban streets and expressways in three different districts in China. Figure \ref{fig: four districts in WHU-Road3D} presents a visual overview of LoD2 road maps of three districts in WHU-Road3D. The collected data exhibits variations in point density, semantic category distribution, and cityscapes among the three districts. 
Table \ref{table: dataset description} lists the statistics of each district in WHU-Road3D. The data collected in District SH covers a road of around 35.153 km, passing through urban streets and a partial expressway. Ten kinds of direction symbols on the ground are gathered. The AS-900HL multi-platform LiDAR measurement system collected the data in District SH. 
The data collected in District WH covers the urban street with a length of 26.18 km using the HiScan-Z 3D laser mobile measurement system. 
In District CM, the LiDAR points were collected by the CHCNAV AS-900HL system, covering a road with a length of 34.185 km.
\par
\begin{figure}[!ht]
\centering
\subfloat[MLS point clouds]{\includegraphics[width=0.4\textwidth] {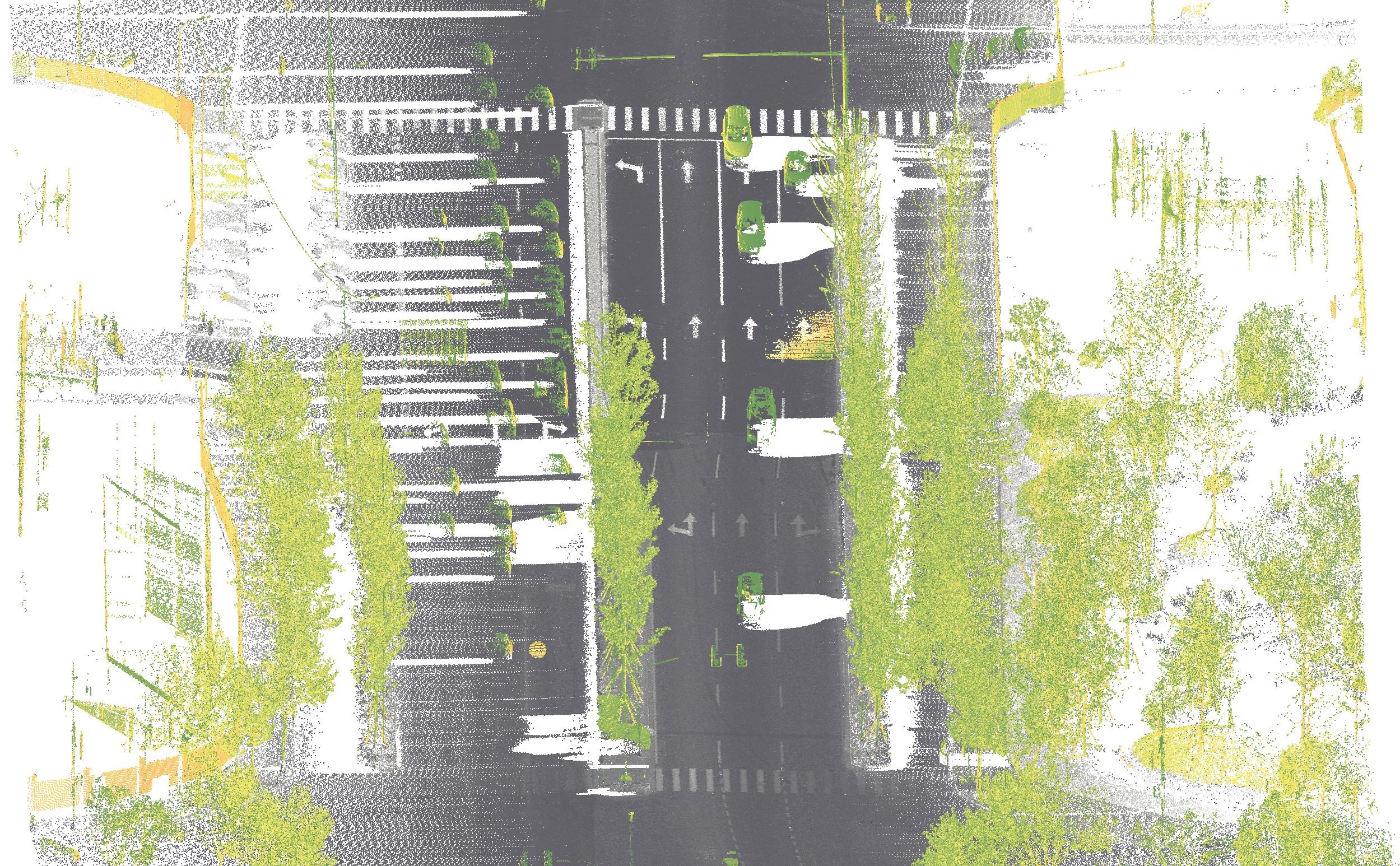}\label{fig_urban_scenario_lidar}}
\vfil
\subfloat[Panoramic image]{\includegraphics[width=0.4\textwidth] {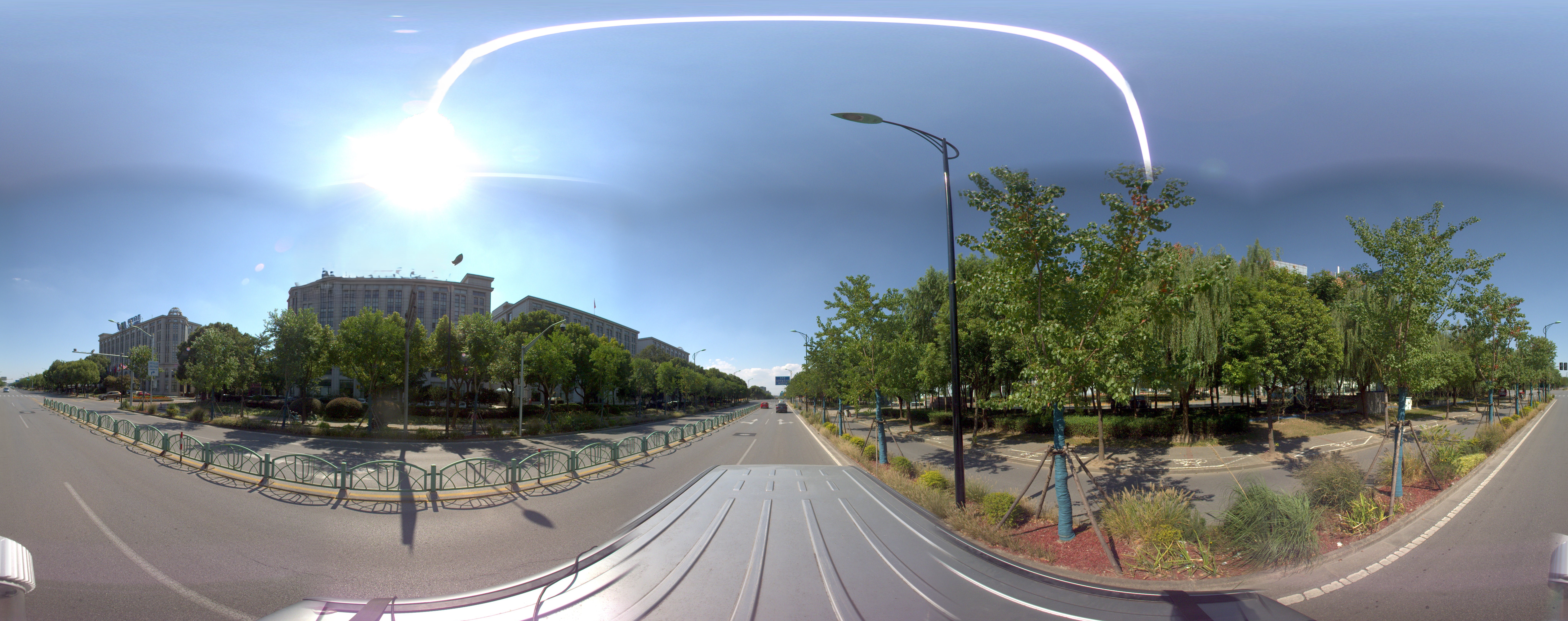}\label{fig_urban_scenario_pano}}
\caption {Typical urban street. Ground points and non-ground point clouds are extracted using CSF \cite{zhang2016easy} and rendered by the intensity in gray and light green, respectively.}
\label{fig: typical urban road scenario}
\end{figure}

The urban streets contain various road infrastructures and dynamic objects, such as poles, traffic signs, traffic lights, trees, bus stops, road intersections, road markings, vehicles, pedestrians, etc. The sizes and shapes of those infrastructures vary between categories and within each category. 
Besides, the interested road surface elements (road boundaries, lane dividers, and direction symbols) suffer significantly from the data incompleteness caused by vehicle occlusion. The busy urban roads usually contain more worn road markings. Figure \ref{fig: typical urban road scenario} displays a typical urban street (panoramic image and MLS point clouds). However, the expressway suffers less from data incompleteness and worn road markings than the urban street. Generally, object extraction methods perform better in expressways than on urban streets.
\par
We performed basic preprocessing operations, including denoising, partitioning, and filtering, on LiDAR points in three districts. The interested categories (road boundaries, lane dividers, crosswalks, and direction symbols) are on the road surface. Therefore, only points within the 30 m buffer radius along the road width and the elevation lower than the nearest trajectory of the scanning vehicle are kept in the datasets.   
\par

\subsection{Dataset annotation}\label{subsec:dataset annotation}
The core component of WHU-Road3D is detailed 3D vectorized annotations of the road surface elements, comprising fifteen categories, including road boundaries, dashed lane dividers, solid lane dividers, crosswalks, stop lines, forward arrows, left turn arrows, right turn arrows, forward and left turn arrows, forward and right turn arrows, turn round arrows, forbidden symbols, rhombus, left and right turn arrows, and curve arrows. Table \ref{table: detection signs statistics} displays the counts of direction symbols across three districts.

\begin{table}[!ht]
    \caption {Category-wise direction symbols' count in three districts.}
    \label{table: detection signs statistics}
	\centering{
	\renewcommand{\arraystretch}{1.3}		
	\resizebox{0.32\textwidth}{!}
    {%
	\begin{tabular}{llll}
	\cline{1-4}
	\multirow{2}{*}{Category}  & \multicolumn{3}{c}{ District }  \\ 
    \cline{2-4}
     & SH & WH & CM   \\ 
	\cline{1-4}
    Forward arrow  & 238      & 263 &  267    \\
    Left arrow  & 141      & 48 &  36    \\
    Right arrow  & 54      & 20 &  18   \\
    Forward \& left arrow  & 37 & 39 &  6     \\
    Forward \& right arrow  & 128  & 62 &  39    \\
    Rhombus  & 9  & 6 &  145    \\
    Stop stripe  & 168      & 51 &  71    \\
    Zebra crossing stripe   & 2447 & 1195 &  1350   \\
    Round arrow   & 10      & 0 &  0    \\
    Forbidden symbol   & 9     & 0 &  0     \\
    Left \& right arrow   & 1 & 0 &  0     \\
    Curve arrow   & 6   & 0 &  0     \\
	\cline{1-4}              
	\end{tabular}%
	}
    }
\end{table}

\begin{figure*}[!htbp]
\centering
\subfloat[Ground point clouds]{\includegraphics[width=0.42\textwidth] {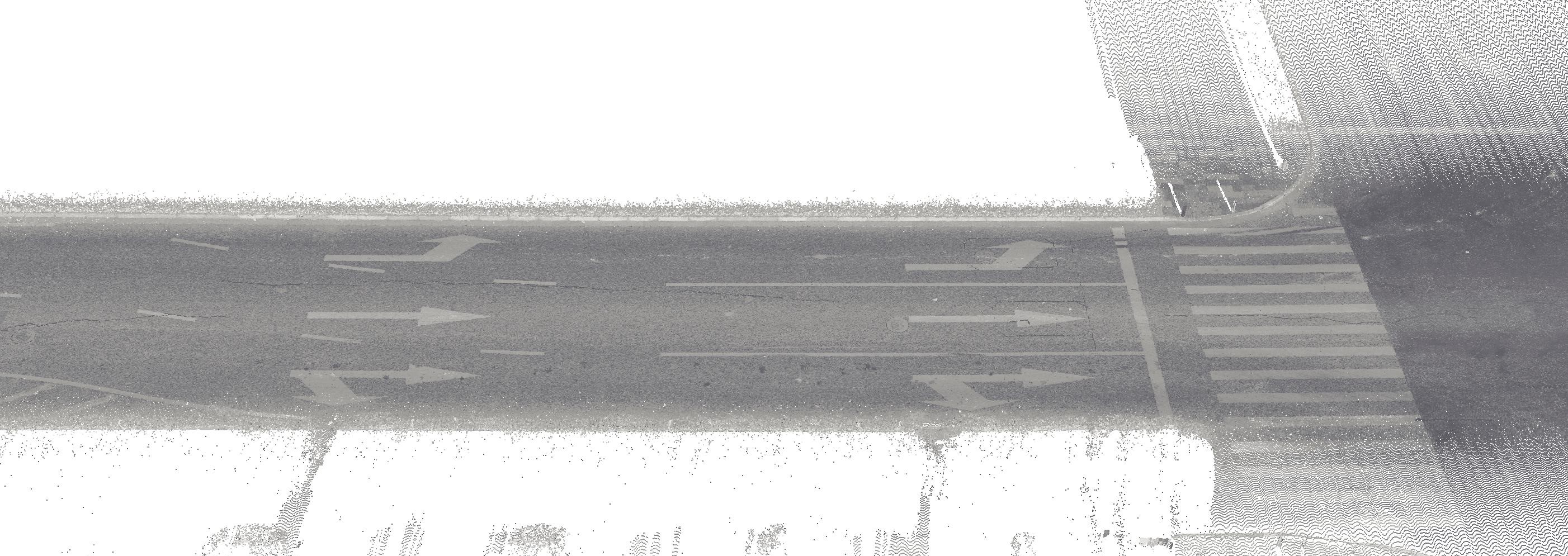}\label{fig_anno-a}}
\hfil
\subfloat[Vectorized annotations]{\includegraphics[width=0.42\textwidth] {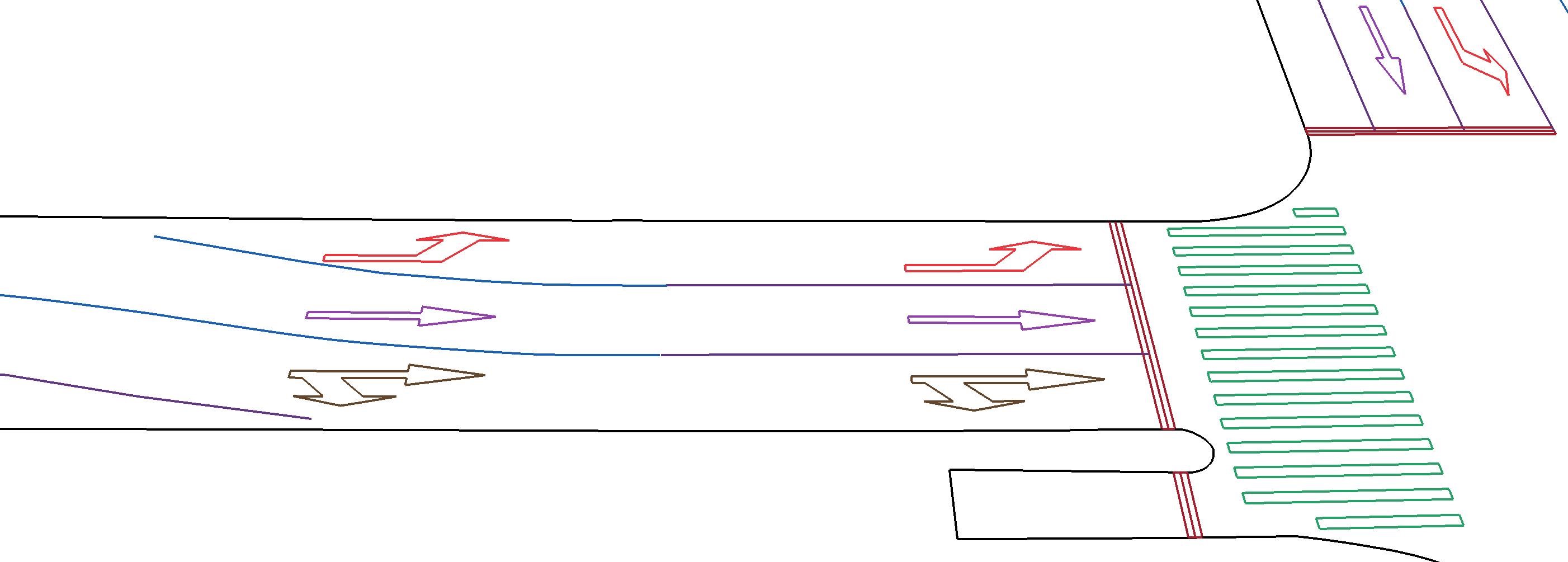}\label{fig_anno-b}}
\vfil
\subfloat[Pointwise semantic annotations]{\includegraphics[width=0.42\textwidth] {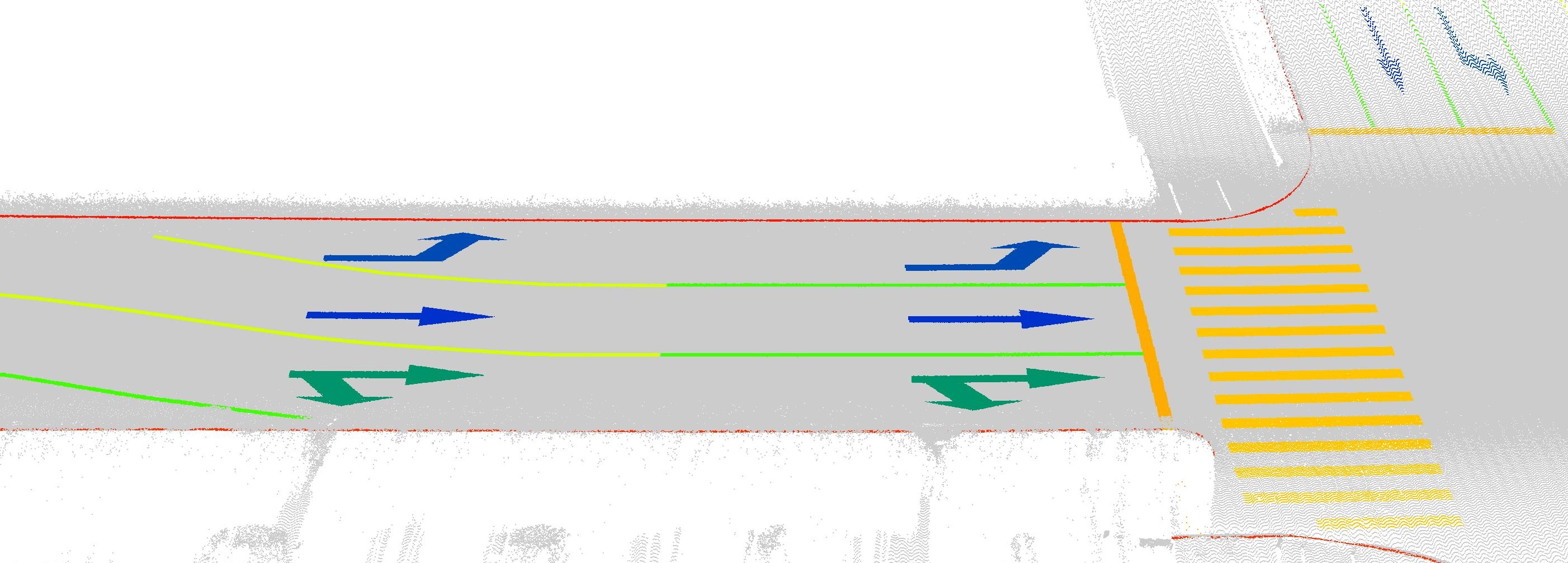}\label{fig_anno-c}}
\hfil
\subfloat[Pointwise instance annotations]{\includegraphics[width=0.42\textwidth] {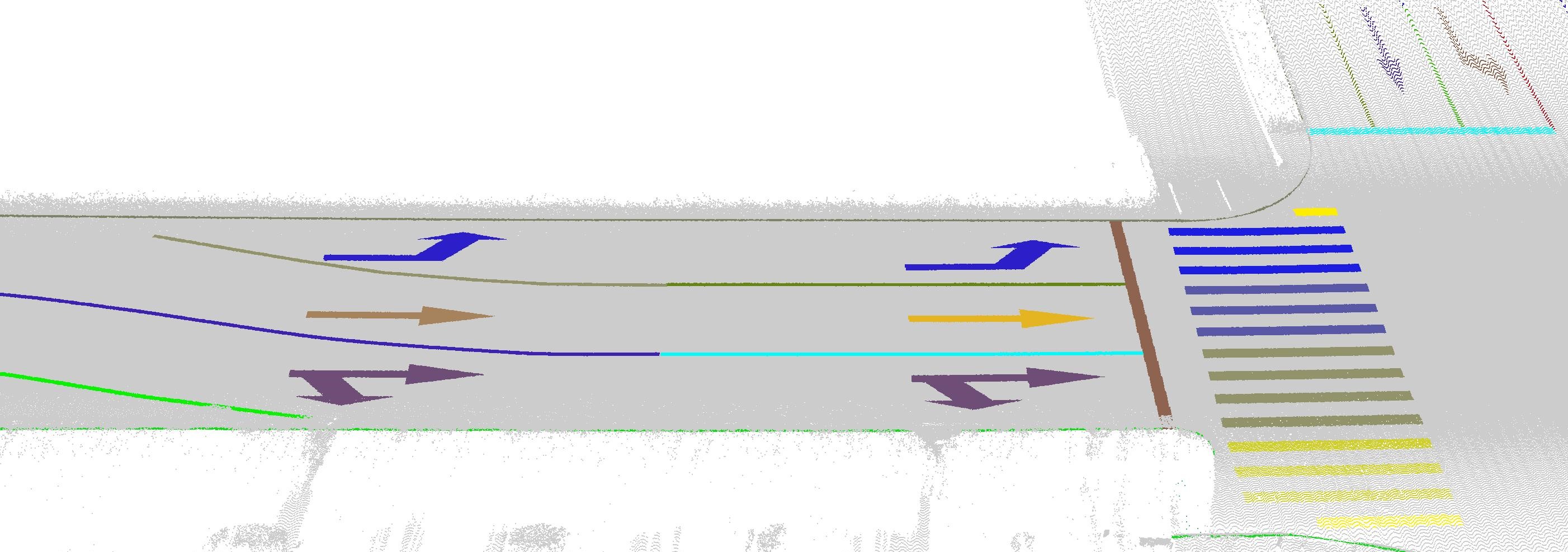}\label{fig_anno-d}}
\caption {Illustration of three kinds of annotations for road surface elements in WHU-Road3D, where (a) shows ground point clouds rendered with descending intensity from light to dark, (b) displays vectorized annotations of road surface elements, (c) and (d) respectively presents pointwise semantic and instance annotations, where semantics and instance IDs are rendered in random colors.}
\label{fig: annotation illustration}
\end{figure*}

\par
We offer pointwise semantic and instance labels, as well as vectorized annotations for road surface elements, as shown in Figure. \ref{fig: annotation illustration}. The vectorized annotation means we directly annotate the road surface elements by compact and structured 3D representations with semantics. We believe that vectorized annotation is more effective than pointwise, which is of preference in downstream applications, such as mapping and spatial analysis \cite{petersGeographicalPointCloud2018}. The following details annotation rules for interested elements on road surfaces. It is worth noting that road elements located at the holes in source point clouds are holistically annotated.
\par
\begin{figure}[!ht]
\centering
\subfloat[Street MLS point clouds]{\includegraphics[width=0.4\textwidth] {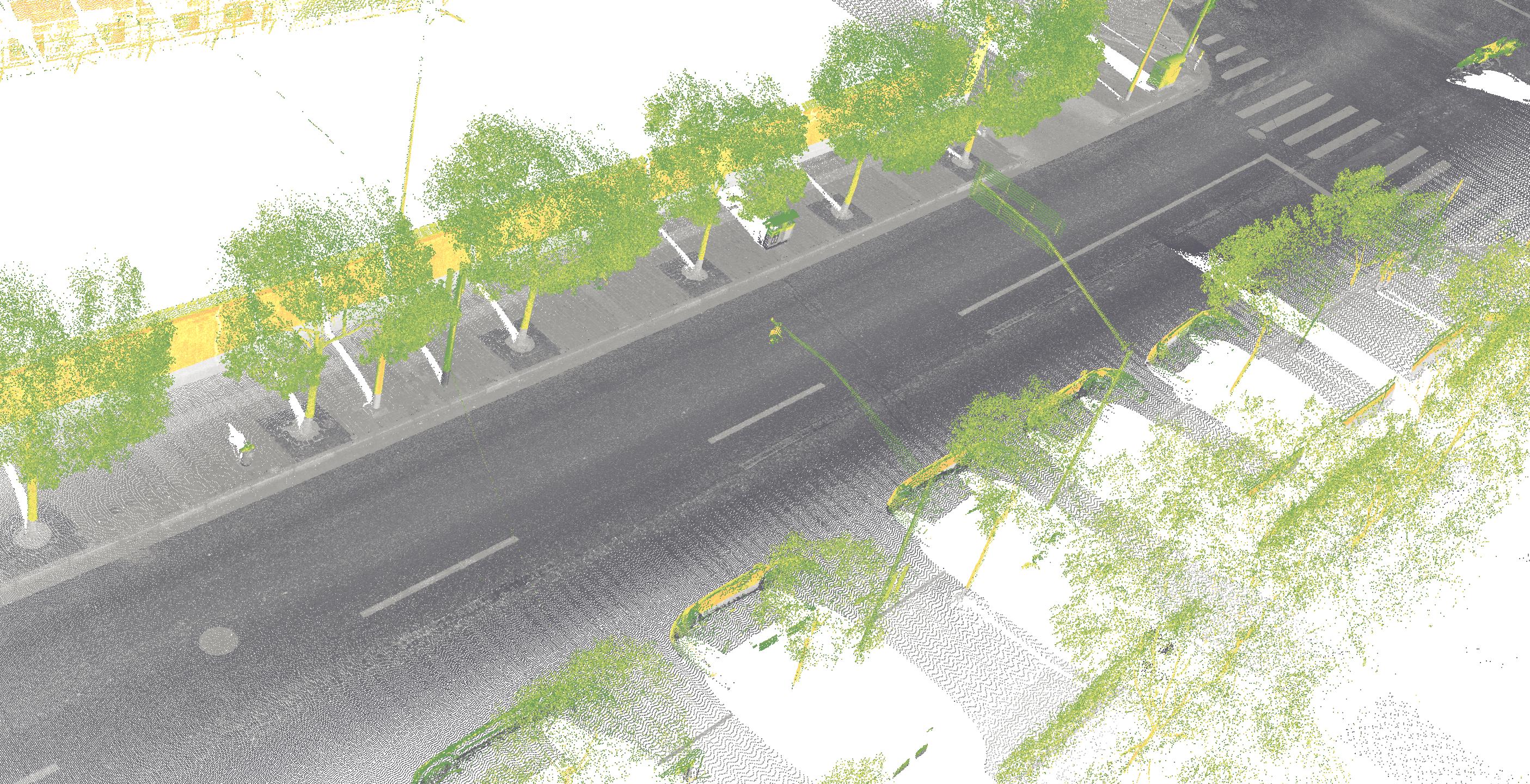}\label{fig_road_boundary_anno-a}}
\vfil
\subfloat[Continuous and complete vectorized annotations]{\includegraphics[width=0.4\textwidth] {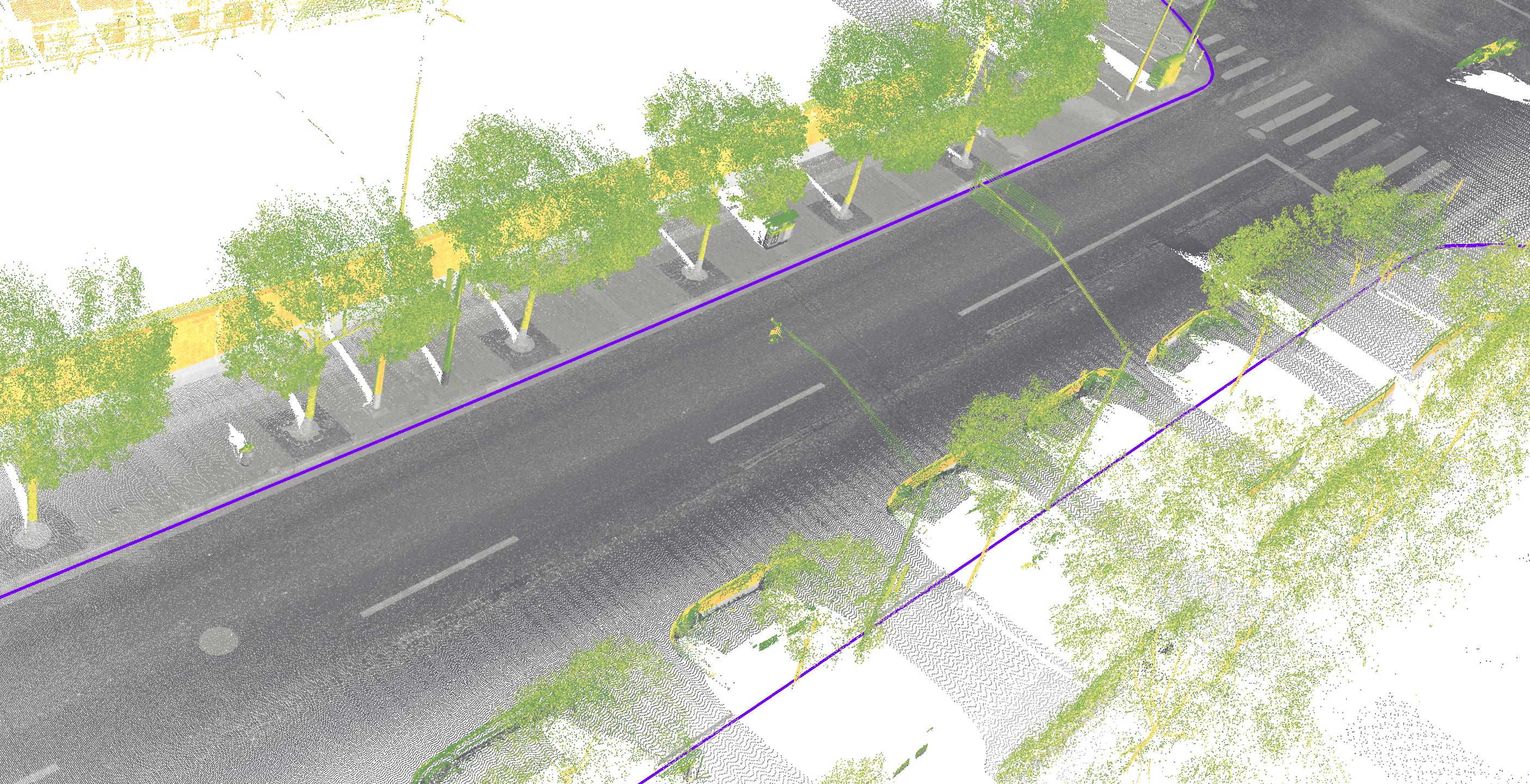}\label{fig_road_boundary_anno-b}}
\caption {Illustration of road boundary annotations, where ground and non-ground point clouds are separately rendered in gray and green, and vectorized road boundary annotation is in dark purple. (a) displays MLS point clouds only. (b) shows complete road boundary annotations even where partial road boundary points are missed in source point clouds.}
\label{fig: road boundary annotation}
\end{figure}

\noindent\textbf{Road boundaries}, also called curbs (stones), are usually used to separate different road functional areas. For instance, typical road boundaries in urban scenarios are displayed in Figure \ref{fig: road boundary annotation}, which are usually artificial road infrastructures with fixed elevation differences to separate the drivable and non-drivable areas for vehicles. We annotate intersections between the vertical curb facade and the horizontal road surface as polylines according to its continuously stretching characteristics. Additionally, we accomplish the complete and continuous annotation in the data hole area, as shown in Figure \ref{fig: road boundary annotation}\subref{fig_road_boundary_anno-b}. 

\begin{figure}[!htbp]
\centering
\subfloat[Point clouds before lane divider annotation]{\includegraphics[width=0.4\textwidth] {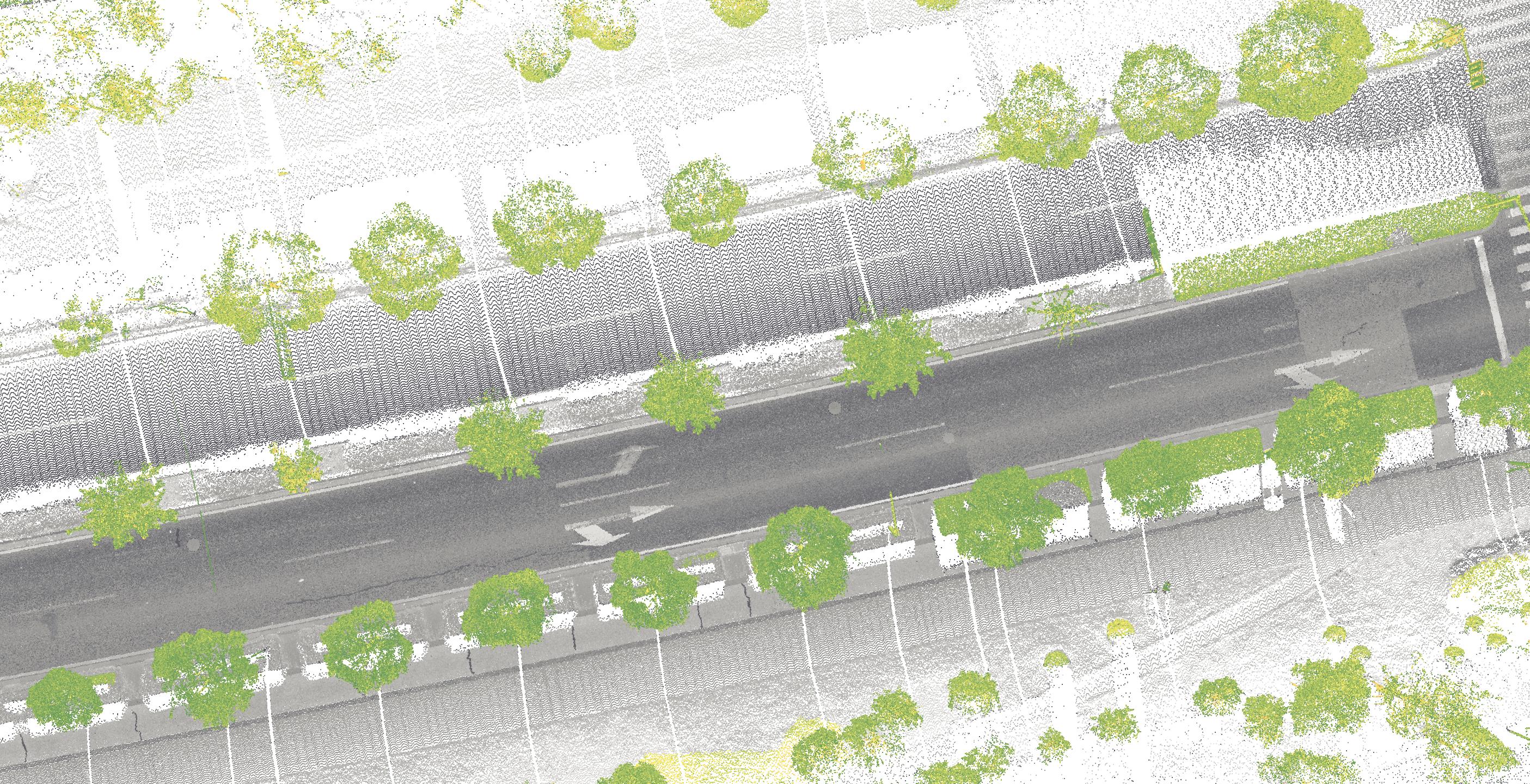}\label{fig_lane_anno-a}}
\vfil
\subfloat[Point clouds overlaying lane divider annotation]{\includegraphics[width=0.4\textwidth] {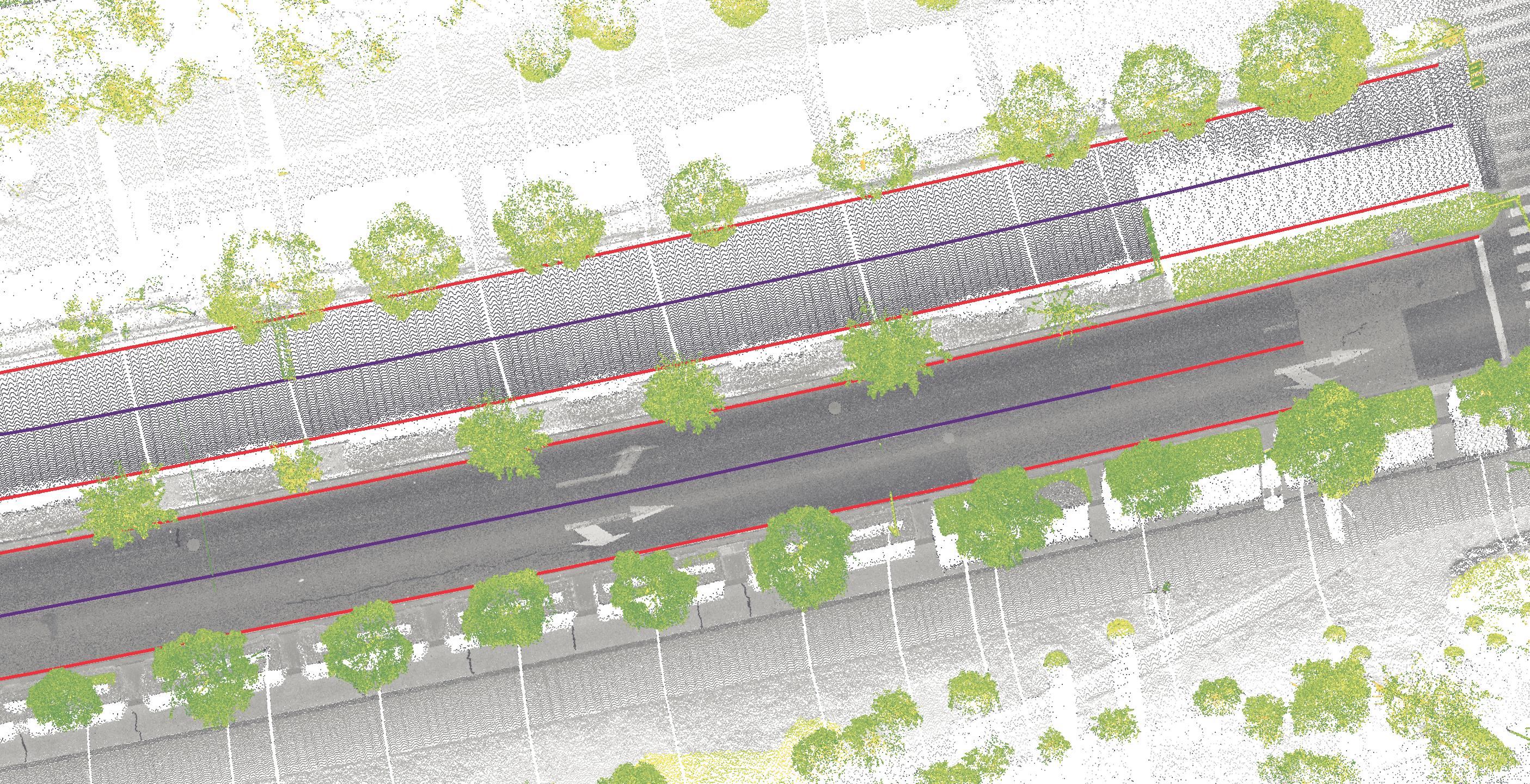}\label{fig_lane_anno-b}}
\caption {Illustration of lane divider annotations, where ground point clouds are rendered in gray, non-ground points are rendered with descending intensity from yellow to green, and vectorized solid and dashed lane dividers are in red and dark purple, respectively. (a) displays point clouds without annotation of lane dividers. (b) shows annotated lane dividers overlaying on point clouds.}
\label{fig: lane divider annotation}
\end{figure}

\noindent\textbf{Lane dividers} outline the drivable lanes for vehicles, which support an orderly driving environment. It can be divided into dashed and solid lane dividers. Dashed lane dividers indicate the connection of the adjacent lanes, while solid lane dividers indicate the disconnection of the adjoining lanes. Specifically, the connection means the topological connection and permits changing lanes between adjacent lanes. Disconnection means the soft ban on changing lanes between adjacent lanes. Meanwhile, road boundaries usually have hard bans on changing lanes for vehicles.
As shown in Fig. \ref{fig: lane divider annotation}, continuous polylines with semantics are employed to annotate lane dividers. 

\begin{figure}[!htbp]
\centering
\subfloat[Intersection point cloud]{\includegraphics[width=0.22\textwidth] {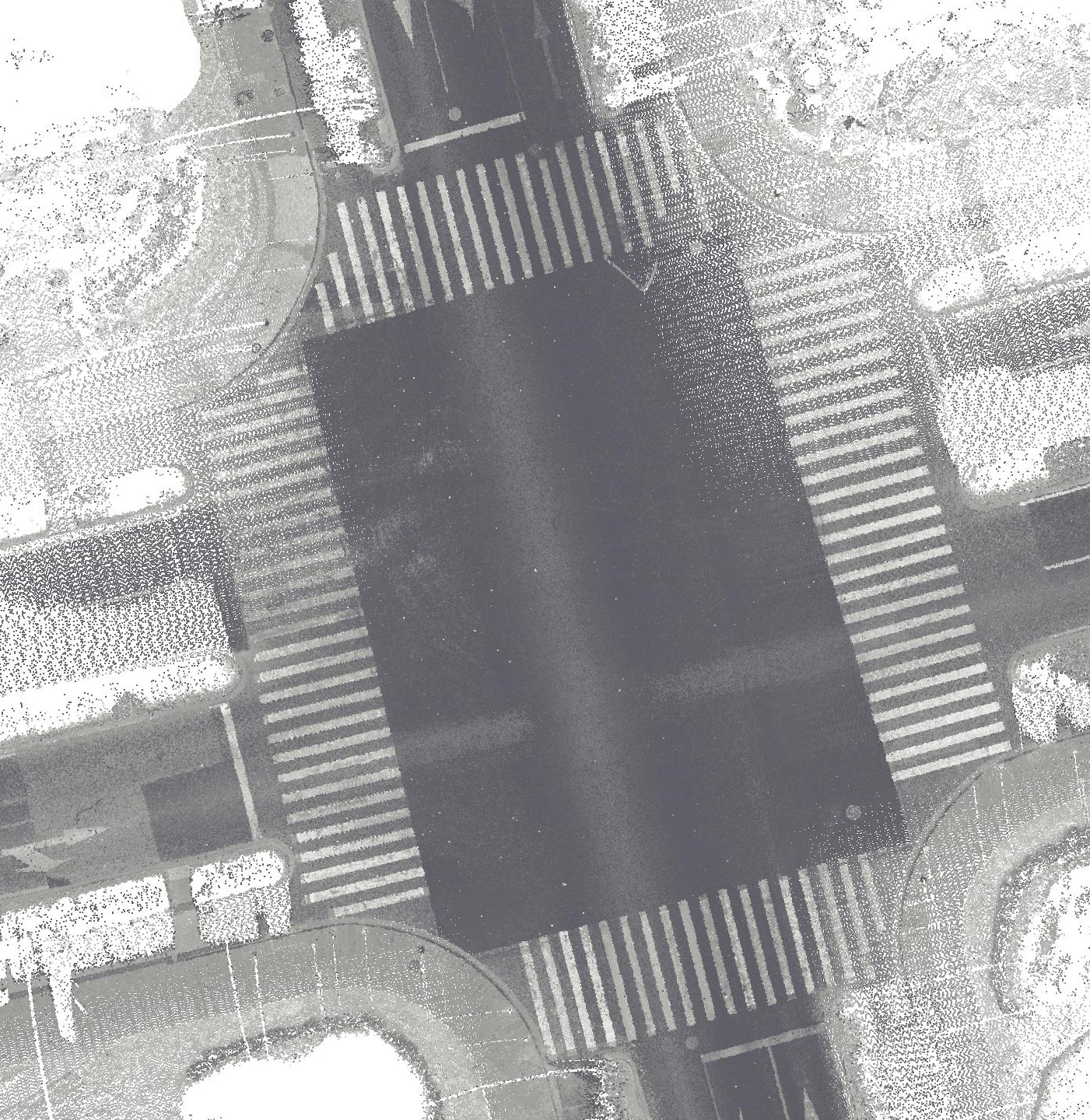}\label{fig_crosswalk_anno-a}}
\hfil
\subfloat[Crosswalks lines]{\includegraphics[width=0.22\textwidth] {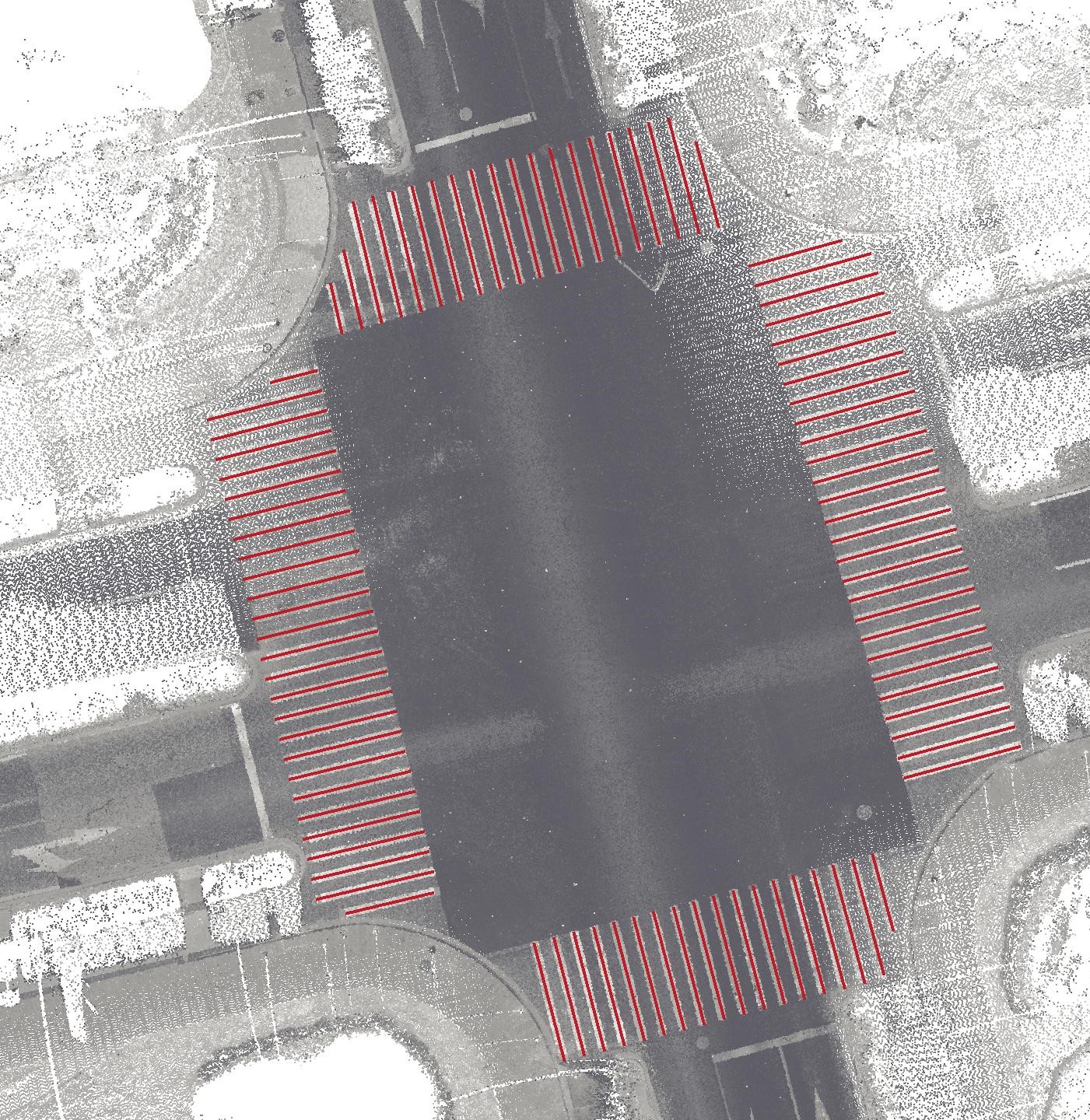}\label{fig_crosswalk_anno-b}}
\vfil
\subfloat[Crosswalks stripes]{\includegraphics[width=0.22\textwidth] {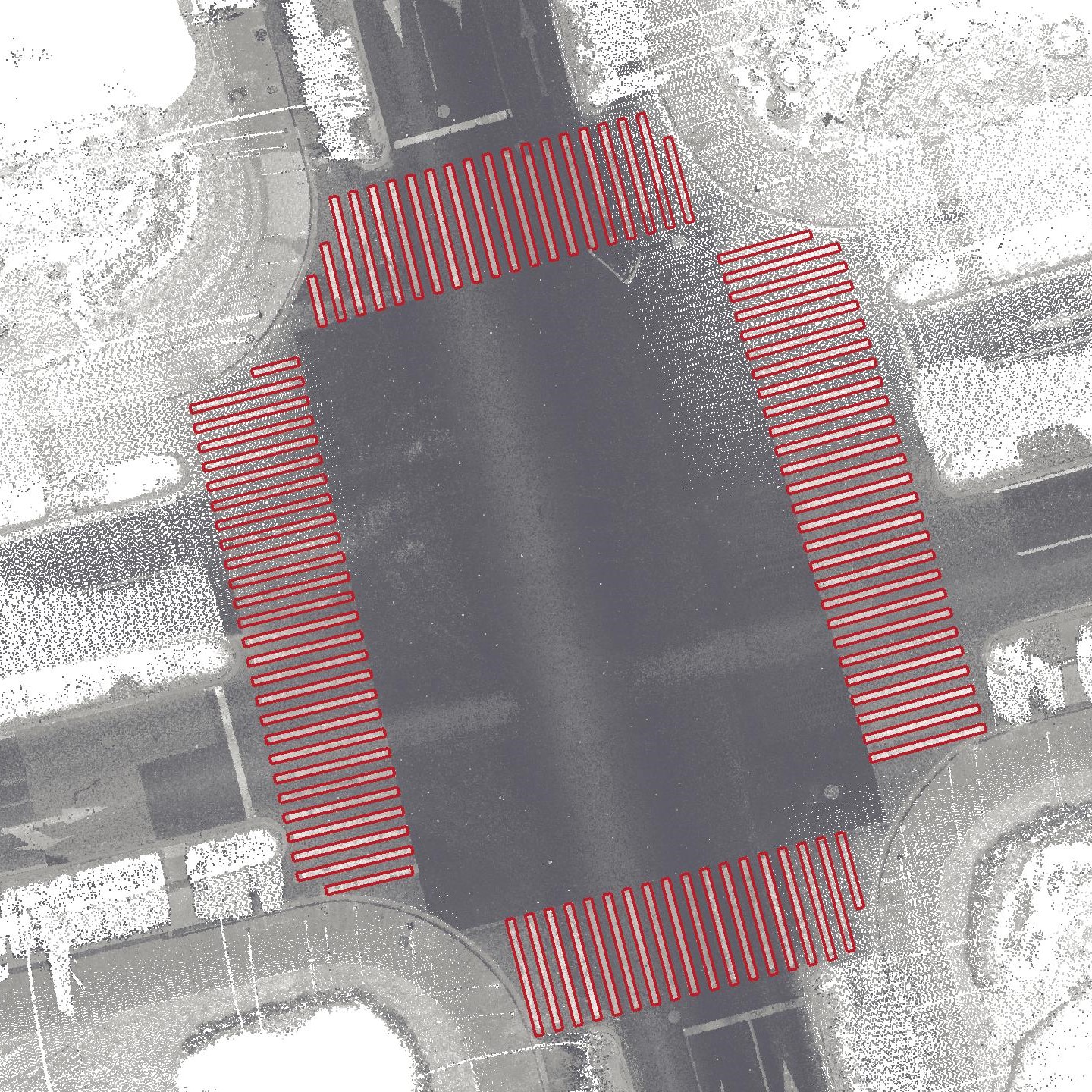}\label{fig_crosswalk_anno-c}}
\hfil
\subfloat[Crosswalks regions]{\includegraphics[width=0.22\textwidth] {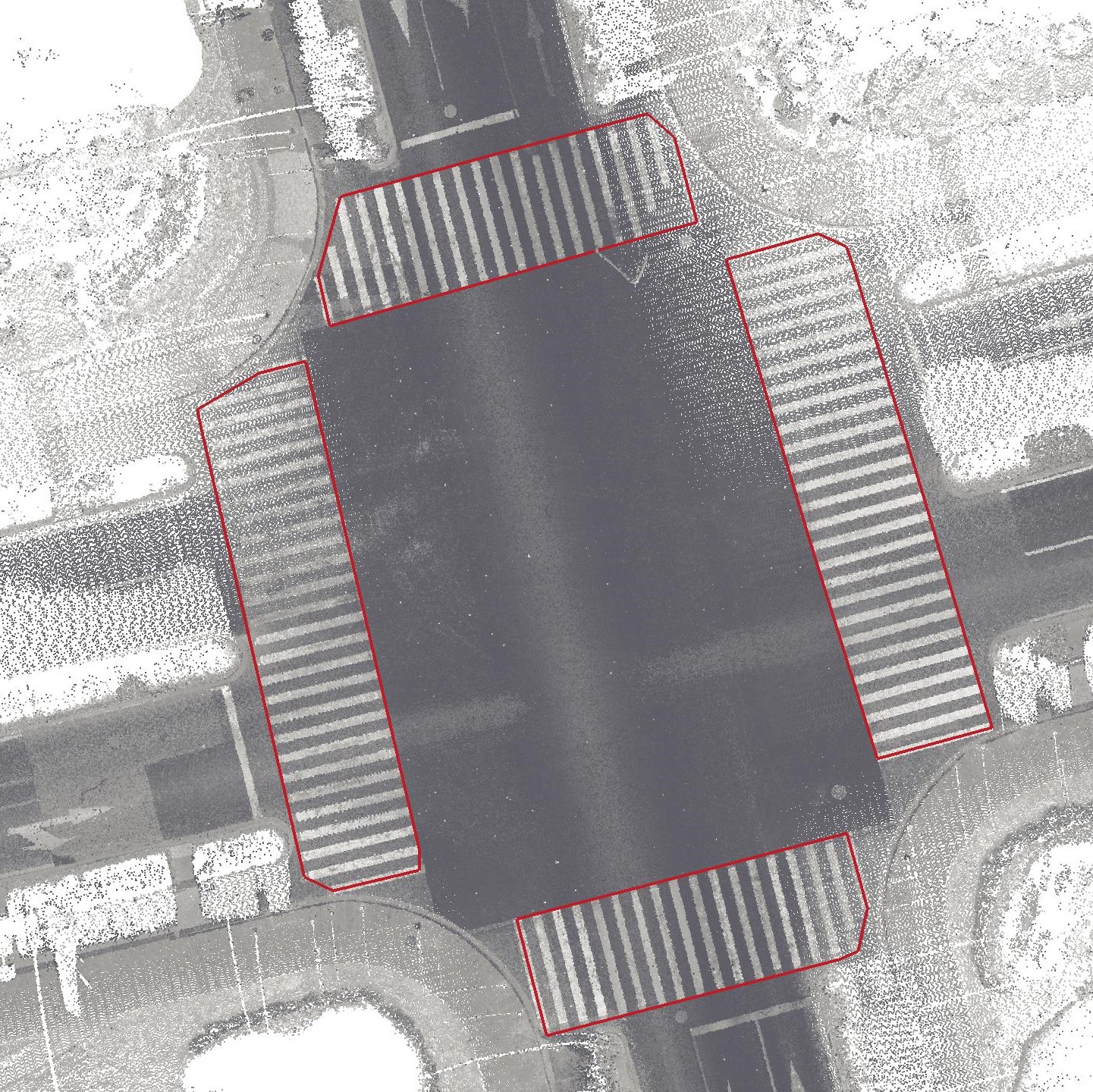}\label{fig_crosswalk_anno-d}}
\caption {Illustration of crosswalk annotations, where ground point clouds are rendered with descending intensity from white to gray, and crosswalks are annotated as lines, stripes, and region outlines.}
\label{fig: crosswalk annotation}
\end{figure}

\noindent\textbf{Crosswalks} are often found at intersections to assist pedestrians to safely cross the road, where pedestrians usually have priority over vehicular traffic. We annotate crosswalks whose appearances are black and white stripes (resembling a zebra). Specifically, we offer three annotation formats for crosswalks, including whole polygon patches, sets of stripes, and bunches of centerlines of stripes, as shown in Figure \ref{fig: crosswalk annotation}. 
\par

\noindent\textbf{Stop lines} are often found in front of crosswalks at intersections, which remind drivers to stop here until the passing signal is on. They are usually in the form of a slender white rectangle. To fit various mapping preferences, we offer annotation formats in polygon centerlines and stipes, similar to crosswalks annotation in Figure \ref{fig: crosswalk annotation}\subref{fig_crosswalk_anno-b} and \subref{fig_crosswalk_anno-c}.  
\par

\begin{figure}[!ht]
\centering{
    \includegraphics[width=0.40\textwidth, height=0.3\textwidth]{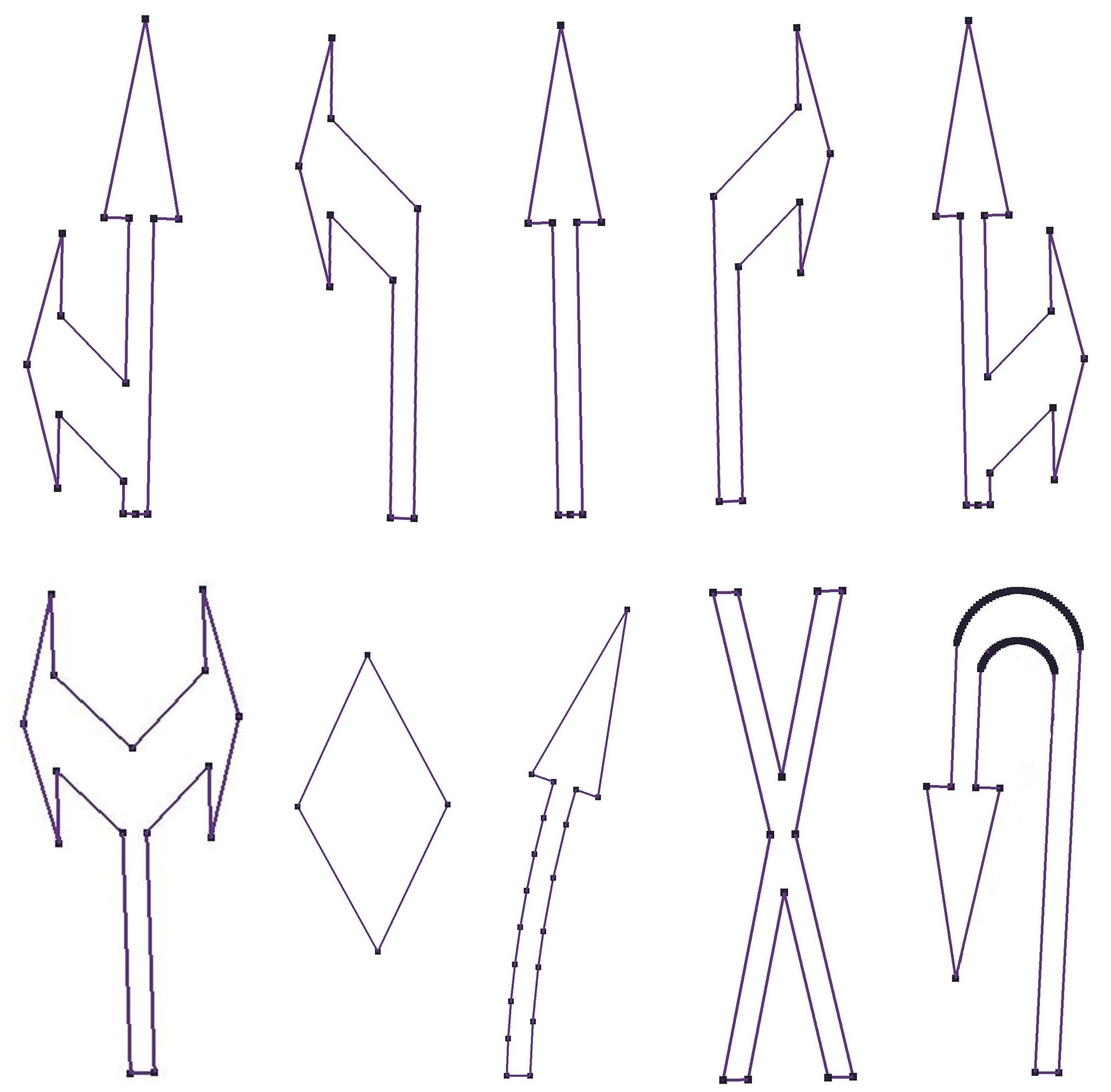}
}
\caption {Ten direction symbols' annotations.}
\label{fig: road marking annotations}
\end{figure}

\noindent\textbf{Direction symbols} are countable arrows on road surfaces, which usually guide the incoming traffic rules. For instance, arrows on the road surface, such as left turn arrows, mean a left-turn permit from the current lane. And rhombuses indicate the coming crosswalks. In WHU-Road3D, we label 10 kinds of typical common symbols on road surfaces, including forward arrows, left-turn arrows, right-turn arrows, forward and left-turn arrows, forward and right-turn arrows, rhombuses, forbidden symbols, round turn arrows, left and right turn arrows, and curve arrows. 
In the WHU-Road3D dataset, direction symbols are annotated using two forms of vectorized representations: oriented bounding boxes and outline contours characterized by a predetermined number of vertices. For instance, seven ordered vertices define the forward arrow compactly. The 3D outlines are annotated for the 10 kinds of interested direction symbols, as shown in Fig. \ref{fig: road marking annotations}. 

\subsection{Advantages of WHU-Road3D}\label{subsec:dataset advantage}
Compared to existing LoD2 road mapping datasets, WHU-Road3D distinguishes itself in three key ways.
\begin{itemize}
\item \textbf{Fine-grained annotations:} In contrast to existing LoD2 road mapping datasets, we carefully annotated fifteen kinds of elements on the road surface. Beyond the standard annotations of lane dividers, road boundaries, and crosswalk outlines, we extensively annotate forward arrows, left-turn arrows, right-turn arrows, forward and left-turn arrows, forward and right-turn arrows, rhombuses, forbidden symbols, round turn arrows, left and right turn arrows, and curve arrows. Moreover, we differentiate between dashed and solid lane dividers. For crosswalks, the dataset offers three distinct types of vectorized annotations, namely, holistic region outlines, sets of individual stripes, and clusters of centerlines, to accommodate diverse needs for map granularity. In addition to vectorized annotations, we include pointwise annotations for all fifteen types of road surface elements. These annotations serve as critical supplementary data for MLS point cloud semantic and instance segmentation, an aspect frequently overlooked in other point cloud segmentation datasets.
The fine-grained annotations developed in this study significantly enrich LoD2 road surface mapping datasets, both in terms of semantic categorization and annotation format.

\item \textbf{Diverse data modalities:} WHU-Road3D encompasses MLS point clouds, panoramic images, vehicle trajectories, and vectorized annotations. Compared to existing LoD2 road mapping datasets, such as nuScenes and Argoverse datasets, MLS point clouds in WHU-Road3D convey evident reflective intensities, enabling direct differentiation of road surface markings. In contrast, most current methodologies for constructing high-definition (HD) maps online predominantly utilize perspective images as the primary data source when utilizing nuScenes and Argoverse datasets. Furthermore, WHU-Road3D includes vehicle trajectories across three distinct districts. Given that road alignments often exhibit consistent elongation patterns, these vehicle trajectories can serve as valuable prior knowledge, potentially enhancing the performance of mapping techniques.
 
\item \textbf{Distinct scenario complexity:} Whereas existing publicly available LoD2 road mapping datasets, such as nuScenes and Argoverse, are sourced from urban environments in the United States and Singapore, WHU-Road3D sets itself apart through its unique cityscape characteristics. The source data and maps in WHU-Road3D are collected from three Chinese cities, encompassing intricate urban streets and expressways. Urban streets in these regions are characterized by a higher density of moving vehicles and data voids, posing substantial challenges for methodologies designed to extract road elements and construct comprehensive maps. In District SH, a notable imbalance in element categories is observed, potentially causing neural networks to prioritize prevalent categories while overlooking less common categories. This category imbalance introduces additional challenges into the extraction and mapping of fine-grained road surface elements.

\end{itemize}

%% file: sections/4-Benchmarks.tex
\section{Benchmark Tasks and Metrics}\label{sec: benchmark tasks and metrics}
WHU-Road3D supports instance segmentation and full-element mapping for road surfaces. This section highlights the extraction and vectorization problem of road surface elements. The fifteen annotated categories' geometric shapes vary. For instance, road boundaries and lane dividers are slender, stretching polylines, while other crosswalks, stop lines, and direction symbols are polygons. Firstly, we provide the evaluation metrics for polylines and polygon objects on road surfaces. Then, we select three baseline methods to offer baseline predictions. Benchmark models are available \href{https://github.com/WHU-USI3DV/WHU-Road3D.git}{here}.

\subsection{Metrics}
\noindent\textbf{Polylines} consists of a set of ordered points. Following OpenLaneV2 \cite{wangOpenLaneV2TopologyReasoning2023}, we adopt precision-recall 
 and the Fréchet Distance ($FD$) to evaluate the predicted polylines. Frechet Distance measures the geometric similarity of two polylines, calculated in equation \ref{equation: frechet distance 1}.
\vspace{-8pt}

\begin{equation}
	D_{Frechet}(P,\hat{P})=\min_{k=1}^{K} ||L_{k}|| ,
\label{equation: frechet distance 1}
\end{equation}
\vspace{-8pt}
\begin{equation}
	L_{k} = \max_{i=1,...,m} D(v_{a_{i}}, \hat{v}_{b_{i}})
\label{equation: frechet distance 2},
\end{equation}

\noindent where $P$ and $\hat{P}$ are predicted and ground truth polylines, in which $p = \left\{v_{a_{1}}, v_{a_{2}}, ..., v_{a_{m}}\right\}$ and $\hat{p} = \left\{\hat{v}_{b_{1}}, \hat{v}_{b_{2}}, ..., \hat{v}_{b_{m}}\right\}$ are subset ordered points of $P$ and $\hat{P}$, respectively. $K$ is the total possible nondecreasing surjection coupling counts between $P$ and $\hat{P}$. $L_{k}$ is the maximum distance between $v_{a_{i}}$ and $\hat{v_{b_{i}}}$.

Precision ($p$), recall ($r$), and F1-score ($f1$) are other standard explicit evaluation metrics calculated by the equation \ref{equation: evaluation pr}. When precision, recall, and F1-score are employed to evaluate the polyline prediction performance, predicted and ground truth polylines require interpolations with a fixed interval. Additionally, the interpolation interval is generally smaller than the buffer radius.
\begin{equation}
    \begin{array}{cc}
         p &= \frac{TP}{TP+FP}, \\ [6pt]
         r &= \frac{TP}{TP+FN},\\ [6pt]
         f1 &= \frac{2\times{p}\times{r}}{p+r},
    \end{array}
    \label{equation: evaluation pr}
\end{equation}
where $TP$ is the number of sampled points on predicted polylines that fall within the buffer zone of sampled points on ground truth polylines; $FP$ is the number of sampled points on predicted polylines that do not intersect with any sampled points' buffer zone on ground truth polylines; $FN$ is the number of sampled points on ground truth polylines without any sampled points on predicted polylines located in their buffer zones. Specifically, once sampled points on ground truth polylines are matched with sampled points on predicted polylines, those points are no longer allowed to match with other sampled points on predicted polylines. On the other hand, researchers, such as \cite{MapTR} and \cite{chen2024maptracker}, take $TP$ as the number of predicted polylines with the Chamfer distance to the ground truth smaller than the given threshold. So dose $FP$ and $FN$.

\noindent\textbf{Polygons} are closed shapes confined by a set of ordered vertices. Precision, recall, and F1-score can be used to evaluate the performance under a fixed IoU threshold. For instance, ${TP}$ indicates the extracted road surface element intersecting with the corresponding ground truth. Additionally, intersection over union (IoU) is usually adopted to evaluate the prediction performance, calculated via equation \ref{equation: iou}.  $IoU$ metric indicates the mapping performance.

\begin{equation}
	IoU = \frac{|P \cap \hat{P|}}{|P \cup \hat{P}|}
\label{equation: iou},
\end{equation}
where $P$ and $ \hat{P}$ are predicted and ground truth polygons.

\subsection{Selection of baselines}\label{subsec: selection of baselines}
In this section, we highlight vectorized mapping of road surfaces, which mainly focuses on lane dividers, road boundaries, crosswalks, stop lines, and direction symbols on road surfaces. 
We analyze the characteristics of each kind of object. Road boundaries are far from road centerlines, distinguished by their height and textural difference from the surroundings. Lane dividers vary in visual patterns, including colors, widths, and continuities. Crosswalks, stop lines, and direction symbols are distinguished by their specific shapes and textures. Road boundaries and lane dividers can be represented as polylines with semantics. Direction symbols are represented by polygons with semantics.
\par
Three methods, WHU-Lane~\cite{miBenchmarkApproachDataset2024}, WHU-RoadMark~\cite{miTwostageApproachRoad2021a}, and MapTR~\cite{MapTR} are adopted to showcase the WHU-Road3D's potential in detailed road surface perception and mapping.
WHU-Lane takes MLS point clouds as the source data to map large-scale dashed and solid lane dividers, similar to WHU-Road3D. Then, publicly available road surface element extraction and mapping methods from MLS point clouds are rare. Therefore, we follow the two-stage approach for road marking extraction and reconstruction ~\cite{miTwostageApproachRoad2021a} and slightly modify the pipeline. This method extracts direction symbols, crosswalks, and stop lines, except road boundaries and lane dividers. To further evaluate the online vectorized HD map construction method on the WHU-Road3D, we adopt the milestone method MapTR based on a unified permutation-equivalent model. MapTR is designed to extract road boundaries, lane dividers, and crosswalks.

\subsection{Experimental setting}\label{subsec: experimental setting}
We split WHU-Road3D into training and testing sets according to the scenario types and object categories. The training set comprises data from Districts SH and CM, while the testing set includes data from District WH. The experimental configurations for each benchmark method are as follows:
\par
\noindent\textbf{WHU-Lane.}
For lane divider mapping, we employ the open-source method WHU-Lane~\cite{miBenchmarkApproachDataset2024} on the WHU-Road3D dataset. In a given road scenario, the model is utilized to predict lane dividers in a vectorized map format. The training data consists of point clouds from Districts SH and CM, while the testing data is sourced from District WH. The MLS point clouds are segmented into 50 m × 25 m patches along the vehicle’s trajectory. The 50 m length is aligned with the direction of travel, and the 25 m width is perpendicular to it. Following the WHU-Lane methodology, the point cloud of each patch is projected into a Bird’s-Eye-View (BEV) image, which serves as the input to the network. Finally, the predicted dashed and solid lane dividers from each patch are merged to construct the global map.

\par
\noindent\textbf{Symbolic Road Marking Mapping.}
Following~\cite{miTwostageApproachRoad2021a}, we formulate the task as a road marking pose estimation problem, where the goal is to determine a 6-DoF (Degrees of Freedom) pose for each marking instance, defined by its location ${(x, y, z)}$, orientation ${(\phi, \theta)}$, and a scale $(s)$. To tackle this problem, we establish a baseline using Oriented R-CNN~\cite{xieOrientedRCNNObject2021, xieOrientedRCNN2024}, an object detection model built upon the MMDetection toolbox~\cite{chenMMDetectionOpenMMLab2019}. We augment the model with an additional binary classification head to enable end-to-end pose estimation on BEV images. This modified model predicts the center $(x, y)$, the width $w$, the $height$, the orientation angle $\theta \in [0, \pi)$ of the oriented bounding box, and a binary flag to resolve its forward direction. 
The model takes BEV images generated from MLS point clouds as input. These BEV images encode three features into the R, G, and B channels, respectively: lowest point elevation, point intensity, and distance to the nearest trajectory point. Ultimately, the framework extracts 12 categories of symbolic road markings, encompassing forward arrows, left turn arrows, right turn arrows, forward and left turn arrows, forward and right turn arrows, turn round arrows, forbidden symbols, rhombus, left and right turn arrows, curve arrows, stop lines, crosswalk strips.

\par
\noindent\textbf{MapTR.}
It is widely recognized as a milestone in HD map construction, owing to its unified formulation and robust performance. We trained the MapTR-tiny model on data from Districts SH and CM, focusing on four categories of linear road markings: dashed and solid lane dividers, road boundaries, and crosswalk contours. Following WHU-Lane \cite{miBenchmarkApproachDataset2024}, the model’s input consists of BEV images generated from LiDAR point cloud. The model was trained for 150 epochs, after which its performance was quantitatively  and qualitatively evaluated on the test set.

\subsection{Experimental evaluation}\label{subsec: experimental evaluation}

\begin{table*}[!htbp]
    \caption {Quantitative results for linear road marking mapping on the WHU-Road3D test dataset, obtained from MapTR~\cite{MapTR} and WHU-Lane~\cite{miBenchmarkApproachDataset2024}. All metrics are calculated using a buffer tolerance of 0.3 meters.}
    \label{table: lane divider result}
	\centering{
    {%
	\begin{tabular}{llllll}
	\cline{1-6}
	Method & Category &  Precision $\uparrow$ & Recall $\uparrow$ & F1 $\uparrow$ & $FD$ $\downarrow$\\
	\cline{1-6}
    \multirow{4}{*}{MapTR~\cite{MapTR}} & Solid lane divider & 0.731 & 0.723 & 0.727 & 1.738 \\
    & Dashed lane divider & 0.752 & 0.795 & 0.773 & 1.763\\
    & Road boundary & 0.397& 0.584& 0.473 & 3.356 \\
    & Crosswalk outline & 0.364 & 0.244 & 0.292 & 4.232 \\
        \cline{1-6} 
	\multirow{2}{*}{WHU-lane~\cite{miBenchmarkApproachDataset2024}} & Solid lane divider  &  0.821 & 0.783 & 0.801 & 1.515  \\
    & Dashed lane divider  &  0.794 & 0.895 & 0.842 & 2.214   \\  
	\cline{1-6}              
	\end{tabular}%
	}
    }
\end{table*}

\begin{figure*}[!ht]
\centering{
    \includegraphics[width=0.7\textwidth]{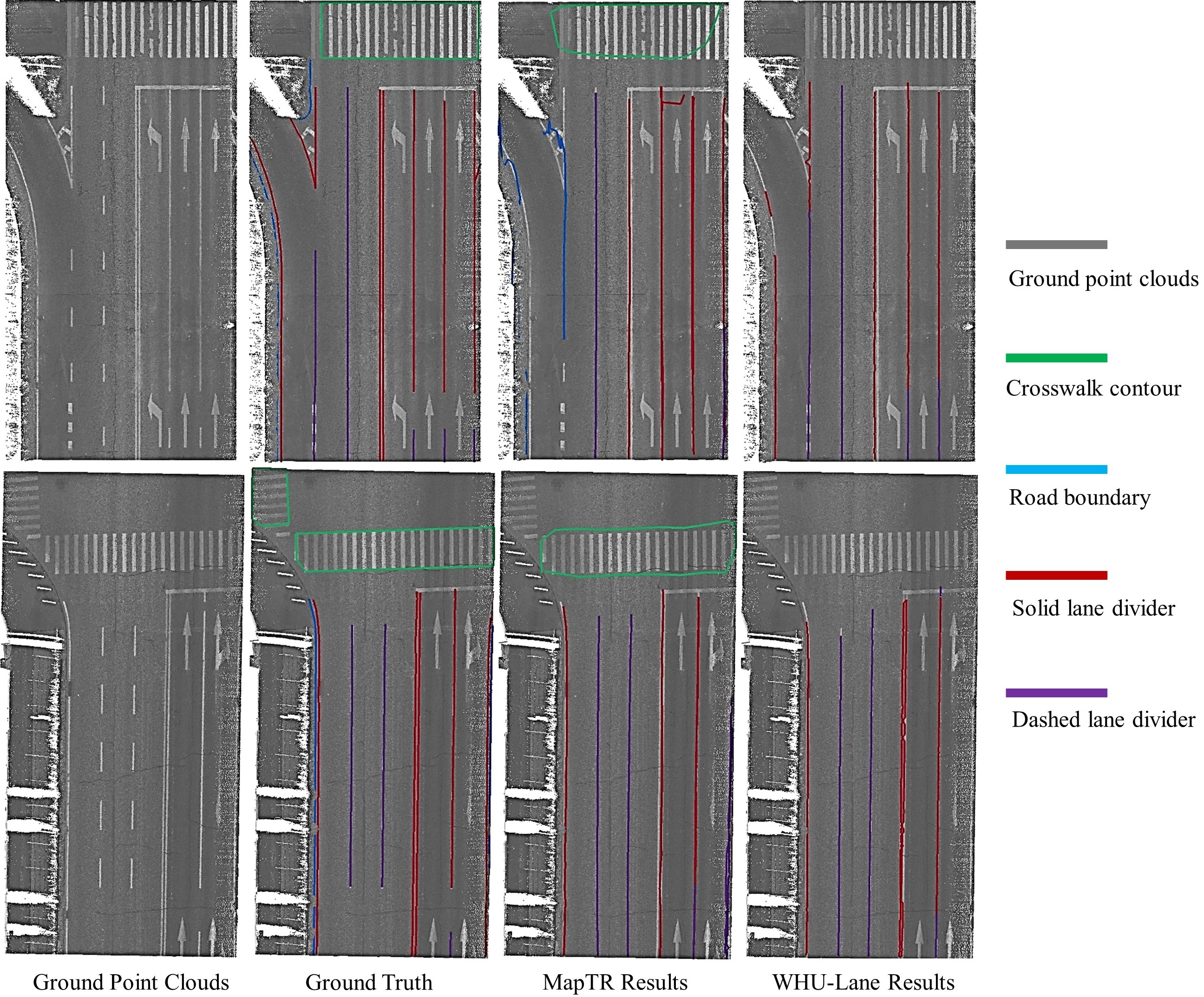}
}
\caption {Visualization results for linear road marking mapping. The first and second columns are ground points rendered with intensity and ground truth annotations of linear road markings (dashed and solid lane dividers, road boundaries, and crosswalk contours), respectively. The third and forth columns are the corresponding predictions by MapTR~\cite{MapTR} and WHU-Lane~\cite{miBenchmarkApproachDataset2024}.}
\label{fig: baseline results for linear road markings}
\end{figure*}

\begin{table*}[!htbp]
    \caption {Quantitative performance of our modified one-stage framework for symbolic road marking mapping on the WHU-Road3D test dataset. Specifically, TP represents the count of predicted symbolic markings that intersects a ground-truth instance. $mIoU$ is the average of the per-category $IoU$ scores.}
    \label{table: direction signs result}
	\centering{
	\renewcommand{\arraystretch}{1.3}		
	\resizebox{0.65\textwidth}{!}
    {%
	\begin{tabular}{lllllllll}
	\cline{1-9}
	Category &  TP &  FP & FN & GT & Precision $\uparrow$ & Recall $\uparrow$ & F1 $\uparrow$& $mIoU$ $\uparrow$\\
	\cline{1-9}
	Forward arrow  & 251      & 25 &  12  &  263 & 0.909 & 0.954 & 0.931 & 0.623 \\
    Left arrow  & 44      & 3 &  4 &  48 & 0.936& 0.917& 0.926 &  0.725  \\
    Right arrow  & 18      & 3 &  2  &  20 & 0.857 & 0.900 & 0.878 &  0.708 \\
    Forward \& left arrow  & 37 & 4 &  2  &  39 & 0.902& 0.949& 0.925 & 0.697  \\
    Forward \& right arrow  & 54 & 3 &  8  &  62 & 0.947& 0.871& 0.907&   0.684\\
    Rhombus  & 6 & 2 &  0  &  6 & 0.750& 1.000& 0.857&  0.437 \\
    Stop stripe  & 32 & 10 &  19  &  51 & 0.762& 0.627& 0.688 &  0.712 \\
    Crosswalk stripe  & 1123 & 44 &  72  &  1195 & 0.962& 0.940& 0.951&  0.741 \\
	\cline{1-9}              
	\end{tabular}%
	}
    }
\end{table*}

\begin{figure}[!ht]
\centering{
    \includegraphics[width=0.38\textwidth]{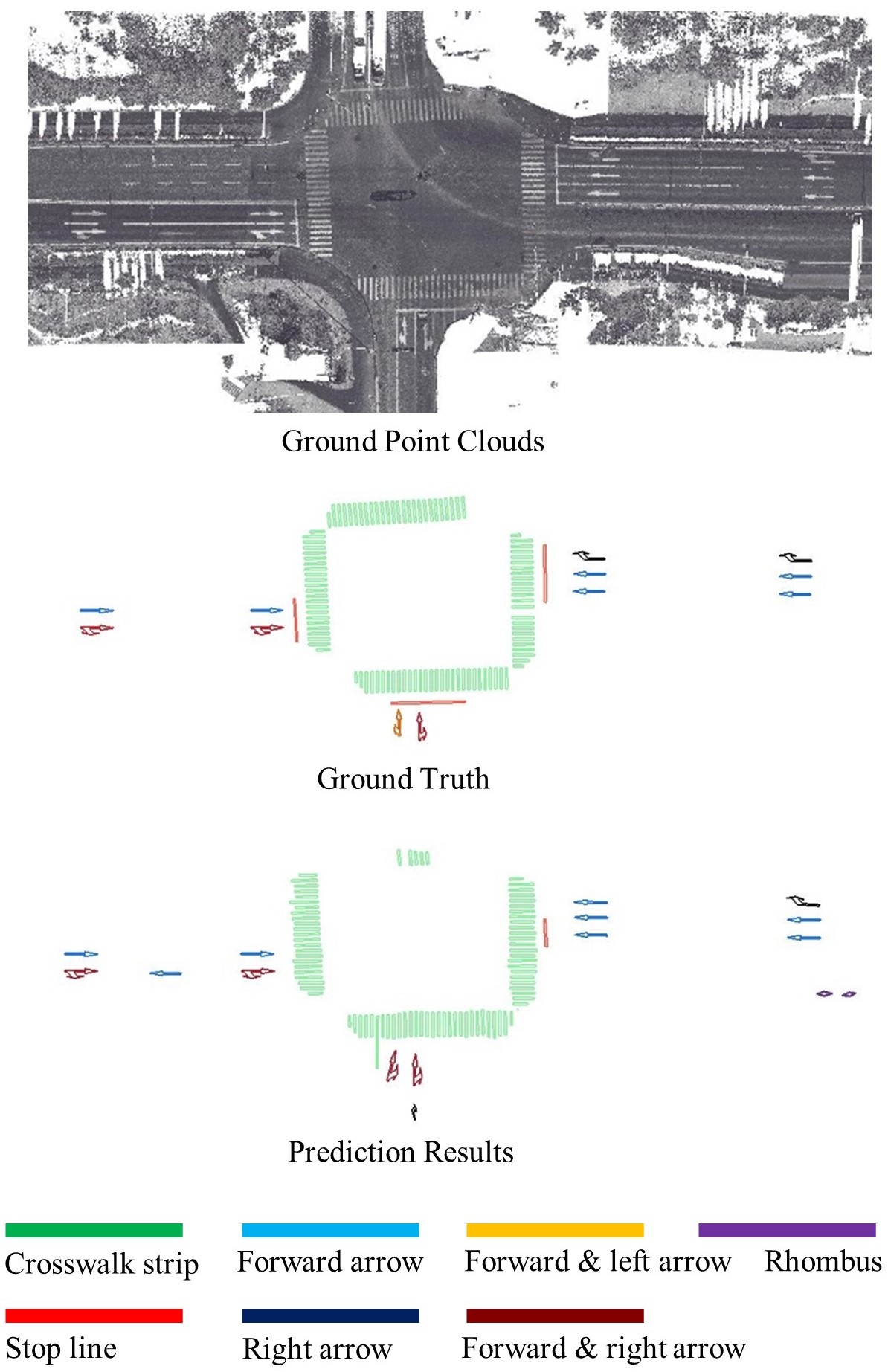}
}
\caption {Visualization results for symbolic road marking mapping. The first and second rows present ground points rendered with intensity and ground truth annotations of symbolic road markings (forward, forward \& left, right, forward \& right arrows, rhombus, crosswalk strips, and stop strips), respectively. The third row shows the corresponding predictions by the modified framework.}
\label{fig: baseline results for symbolic road markings}
\end{figure}

\noindent\textbf{Linear Road Marking Mapping.}
Table \ref{table: lane divider result} presents the quantitative performance of two baseline methods, MapTR~\cite{MapTR} and WHU-Lane~\cite{miBenchmarkApproachDataset2024}, evaluated using precision, recall, F1-score, and Fréchet Distance. MapTR~\cite{MapTR} predicts a broader range of linear markings including lane dividers, road boundaries, and crosswalk contours, while WHU-lane~\cite{miBenchmarkApproachDataset2024} is limited to solid and dashed lane dividers. All metrics were computed with a buffer tolerance of 0.3 m, a stricter criterion compared to the original MapTR evaluation. The MapTR is employed more lenient tolerances of 0.5 m, 1.0 m, and 1.5 m.
According to the quantitative results from MapTR~\cite{MapTR}, it achieves significantly better results on lane dividers and road boundaries than on crosswalks. This disparity is likely because its polyline formulation is better suited to the linear geometry of lane dividers and boundaries. Additionally, the model performs better on dashed lane dividers than solid ones. The outcome is because of the spatial proximity between solid lane dividers and road boundaries.
A key observation across both models is the substantially higher recall for dashed lane dividers compared to solid ones. This is partly because both models are prone to misclassifying lane lines as dashed in dash-to-solid transition regions, where annotations are void. 
Figure~\ref{fig: baseline results for linear road markings} illustrates these situations with mapping results from two representative scenarios.
These findings suggest that the effective aggregation of global features and local details is critical for improving the mapping performance of long-stretching alignments on road surfaces. In addition to more powerful feature encoders, incorporating spatial layout constraints into the model’s formulation holds promise for further performance gains. For instance, modeling the explicit spatial rule that a road boundary typically runs parallel to, and outside of, the nearest solid lane divider could prevent confusion where solid lane dividers run closely parallel to road boundaries.

\noindent\textbf{Symbolic Road Marking Mapping.}
Table \ref{table: direction signs result} presents the quantitative performance for symbolic road marking mapping results of test dataset in terms of precision, recall, F1, and $mIoU$ metrics. Figure~\ref{fig: baseline results for symbolic road markings} visualizes the ground point clouds,  symbolic road markings' ground-truth annotations, and predictions of the modified framework for an urban intersection. The framework’s unified 6-DoF formulation enables the detection and reconstruction of a wide variety of symbolic markings within a single network. Despite its overall strong performance, the framework struggles specifically with stop lines. This challenge arises because the oriented object detection backbone is less effective at localizing geometrically slender objects, such as the rectangular shape of stop lines. Another observed weakness is that the baseline framework erroneously detects candidate symbolic road markings and degrades the quality of the final map. Future work could therefore focus on improving slender object detection to enhance overall mapping accuracy.

%% file: sections/5-FutureDirections.tex
\section{Challenges and future research prospect}\label{Future directions}
\subsection{Full-element road surface mapping}
Roads serve as the vital arteries of urban transportation systems, comprising a network of diverse artificial infrastructures. Currently, popular publicly available datasets for fine-grained road mapping, such as nuScenes~\cite{caesarNuScenesMultimodalDataset2020} and Argoverse~\cite{changArgoverse3DTracking2019, wilsonArgoverse2Next2021}, focus on limited categories of on-surface elements (lane dividers, road boundaries, and crosswalks). These critical datasets have facilitated the development of numerous automated high-definition (HD) map construction methodologies. According to the main concerned objects, most of the proposed methods~\cite{li2022hdmapnet, MapTR, liu2023vectormapnet, chen2024maptracker} conceptualize these elements as ordered sequences of points or segmented curves. However, such formulations prove inadequate for representing direction symbols. Hopefully, researchers will bring up an alternative formulation for comprehensive road surface mapping, leveraging the WHU-Road3D dataset. Moreover, the diverse requirements of downstream tasks necessitate varying levels of detail in road surface maps. To address this, WHU-Road3D offers three distinct annotation types for crosswalks and two for direction symbols, thereby accommodating a broader spectrum of application needs.

\subsection{Spatial layouts assisted mapping}
Road infrastructure typically exhibits a regular distribution and adheres to distinct spatial configurations. For example, lane dividers and road boundaries extend parallel to the direction of the roadway, while crosswalks consist of uniform stripes sharing a consistent orientation. Stop lines, positioned proximate to crosswalks, are oriented perpendicularly to the adjacent zebra stripes. Similarly, direction symbols, consistently located within lanes, align with the principal orientation of the roadway. The effective integration of these spatial layouts is essential for machine learning techniques to understand the interrelationships among various elements on road surfaces, further improving the geometric regularization and topological correctness.
Currently, most methods focus on shared semantic and geometric properties within individual categories while overlooking the spatial layouts across different categories. 
With the valuable resource in the WHU-Road3D dataset, future research could explore the inherent regularities of the built road environment. For instance, simultaneously collected vehicle trajectories can be adopted to support fine-grained road mapping efforts using WHU-Road3D.       

\subsection{Few-shot and open-vocabulary road mapping}
Road boundaries, lane dividers, crosswalks, and direction arrows (forward arrows, left-turn arrows, right-turn arrows, forward and left-turn arrows, etc.) constitute the predominant elements of road surface markings. However, categories with limited samples are susceptible to misidentification, a concern of notable significance. Consequently, researchers may direct their efforts toward the development of few-shot object detection techniques to enhance the mapping accuracy of categories with fewer samples, such as curve arrows, forbidden symbols, and left and right turn arrows. On the other hand, specialized symbols, including bicycle icons and Chinese characters, are absent from datasets such as Whu-Road3D. Visual Language Models (VLMs), trained on millions of image-text pairs \cite{zhang2022dinodetrimproveddenoising, li2023blip, liu2024grounding}, demonstrate robust zero-shot learning and chain-of-thought reasoning\footnote{\url{https://openai.com/index/gpt-4-research/}, \url{https://ai.meta.com/blog/meta-llama-3/}, \url{https://huggingface.co/Qwen}, \url{https://www.deepseek.com/}}. 
Researchers may thus explore the integration of VLMs into fine-grained road mapping to bolster the model's generalization across diverse and challenging scenarios \cite{peng2023kosmos2groundingmultimodallarge, zhang2024mm, WANG2025105995}.   

\subsection{Hierarchical road topology reasoning}
Maps serve strong prior knowledge for vehicles, delivering comprehensive geometric, semantic, and topological road information around and beyond the reach of onboard sensors. Several datasets are released for road mapping concerning road centerlines, road boundaries, lane dividers, etc. However, those datasets do not consider direction symbols on road surfaces, which are vital objects in road surface mapping. 
On the other hand, directional symbols indicate permissible future routes, which are significant in reasoning hierarchical road topology. 
While some studies have explored road topology reasoning via crowd-sourced trajectories and perspective images \cite{hierarchicalroad2022zhang, NEURIPS2024_7116cda4, lv2024t2sgtraffictopologyscene, kalfaoglu2024topomaskv2enhancedinstancemaskbasedformulation} to address road-level and lane-level topology, the resulting topological information lacks regional hierarchical road topology. Regional hierarchical topology necessitates understanding hierarchical relationships among streets, carriageways, and lanes \cite{Zhu01052008}, as well as the connectivity between adjacent lanes. WHU-Road3D maintains regional LiDAR point clouds, panoramic images, and detailed vectorized annotations encompassing nearly all road surface elements. It can support studies on road topology reasoning across various data modalities. Hopefully, using WHU-Road3D, researchers can tackle the challenges of hierarchical road topology, progressing from local scenes to broader regions shortly.         

\subsection{Lane-level change detection and update}
Roads undergo annual construction and reconstruction worldwide, resulting in requests for frequent updates to road maps. Beyond updates at the road centerline level, the lane-level update is crucial. WHU-Road3D encompasses diverse scenarios (urban streets and expressways) in China with accurate LoD2 (lane-level) vectorization maps. It can facilitate road asset management and static layer construction in HD maps. Moreover, we will continuously enrich the WHU-Road3D every four years, incorporating updated panoramic images and vectorized map revisions that account for changes in both geometry and topology. Notably, road map change involves geometry, semantics, and topology. Generally, geometrical and semantic changes often relate to the change in topology. 
It is expected that the realistic WHU-Road3D dataset will advance research into detailed map updates, thereby contributing to the development of accurate and up-to-date road maps.

\subsection{Street scene generation from vectorized maps}
Rational 3D street scene generation is an open problem in virtual reality and autonomous driving systems, which is a promising solution to affordable scene reconstruction and corner case simulation. WHU-Road3D maintains regional, fine-grained 3D road surface layouts, panoramic images, and MLS point clouds, covering urban streets and expressways. Researchers can adopt those data to generate novel view images and scenes \cite{han2024ggsgeneralizablegaussiansplatting}. Apart from the realistic textures of the scenario, the generated 3D scenes should maintain geometric consistency and rational topology. The structural representation of the road surface in WHU-Road3D is in compact vectorization, making it easy to edit the elements' location, appearance, and attributes. For example, a street with novel drivable routes could be generated by removing, adding, or changing the semantics of a lane divider from the dashed to the solid. The explicit constraints in geometry, semantics, and topology are with good flexibility and interpretability. Hopefully, WHU-Road3D could support studies on rational scene generation \cite{gao2024magicdrive3d, gao2023magicdrive, deng2024streetscapes}, using explicit road layouts (including lane dividers, direction symbols, stop lines, and zebra crossings) as constraints.    

%% file: sections/6-Conclusion.tex
\section{Conclusion}\label{section: conclusion}
We have introduced a new dataset, WHU-Road3D, for fine-grained extraction and mapping of road surface elements. WHU-Road3D stands out due to its unique data modalities, rich annotation content, and complex coverage scenarios. Additionally, we provide baseline methods for mapping dashed and solid lane dividers, road boundaries, crosswalks, stop lines, and direction symbols. Hopefully, this newly released multi-modal dataset will create new opportunities for the full-element road surface mapping, hierarchical road topology reasoning, lane-level change detection, and street scene generation.